\pdfoutput=1

\documentclass[11pt]{article}
\usepackage[table,xcdraw]{xcolor}

\usepackage[final]{acl}

\usepackage{times}
\usepackage{latexsym}

\usepackage[T1]{fontenc}

\usepackage[utf8]{inputenc}

\usepackage{siunitx}

\usepackage{graphicx}

\usepackage{microtype}

\usepackage{inconsolata}

\usepackage{graphicx}
\usepackage{subcaption}

\usepackage{amsmath}
\usepackage{booktabs}

\usepackage{multirow}

\usepackage{tcolorbox}

\usepackage{longtable} 
\usepackage{graphicx} 
\usepackage{array} 
\usepackage{lscape} 

\usepackage[normalem]{ulem}
\useunder{\uline}{\ul}{}

\usepackage{hyperref}

\title{Burn \textit{After} Reading: Do Multimodal Large Language Models Truly Capture Order of Events in Image Sequences?   }

\author{Yingjin Song, Yupei Du, Denis Paperno and Albert Gatt \\
  Utrecht University, Utrecht, the Netherlands \\
  \texttt{\{y.song5, y.du, d.paperno, a.gatt\}@uu.nl} }

\begin{document}
\maketitle
\begin{abstract}
This paper introduces the TempVS benchmark, which focuses on temporal grounding and reasoning capabilities of Multimodal Large Language Models (MLLMs) in image sequences. 
TempVS consists of three main tests (i.e., event relation inference, sentence ordering and image ordering),  each accompanied with a basic grounding test.
TempVS requires MLLMs to rely on both visual and linguistic modalities to understand the temporal order of events.
We evaluate \textbf{\textit{38}} state-of-the-art MLLMs, demonstrating that models struggle to solve TempVS, with a substantial performance gap compared to human capabilities. We also provide fine-grained insights that suggest promising directions for future research.
Our TempVS benchmark data and code are available at \url{https://github.com/yjsong22/TempVS}.


\end{abstract}

\begin{figure*}[h]
    \centering
    \begin{subfigure}{0.495\linewidth}
        \centering
    \includegraphics[width=\linewidth, trim=0 225 0 0, clip]{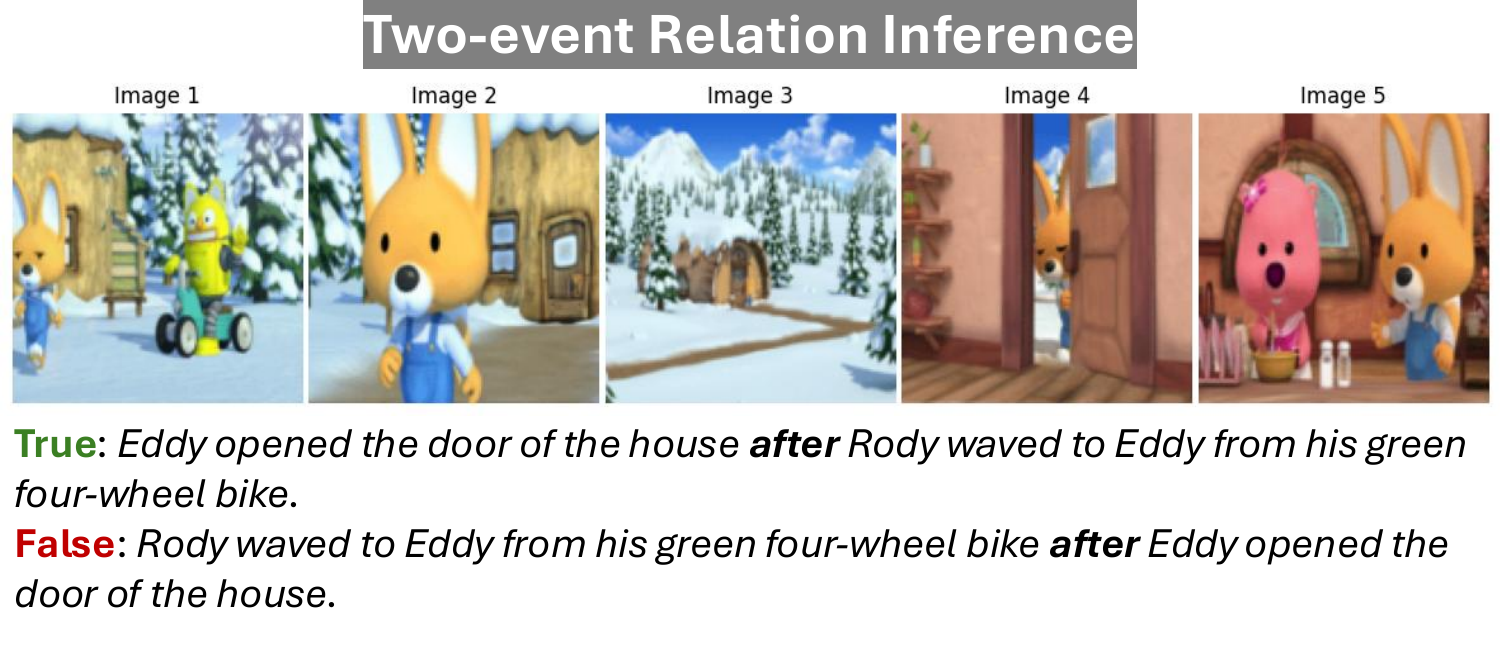}
    \end{subfigure}
    \begin{subfigure}{0.495\linewidth}
        \centering
       \includegraphics[width=\linewidth, trim=0 225 0 0, clip]{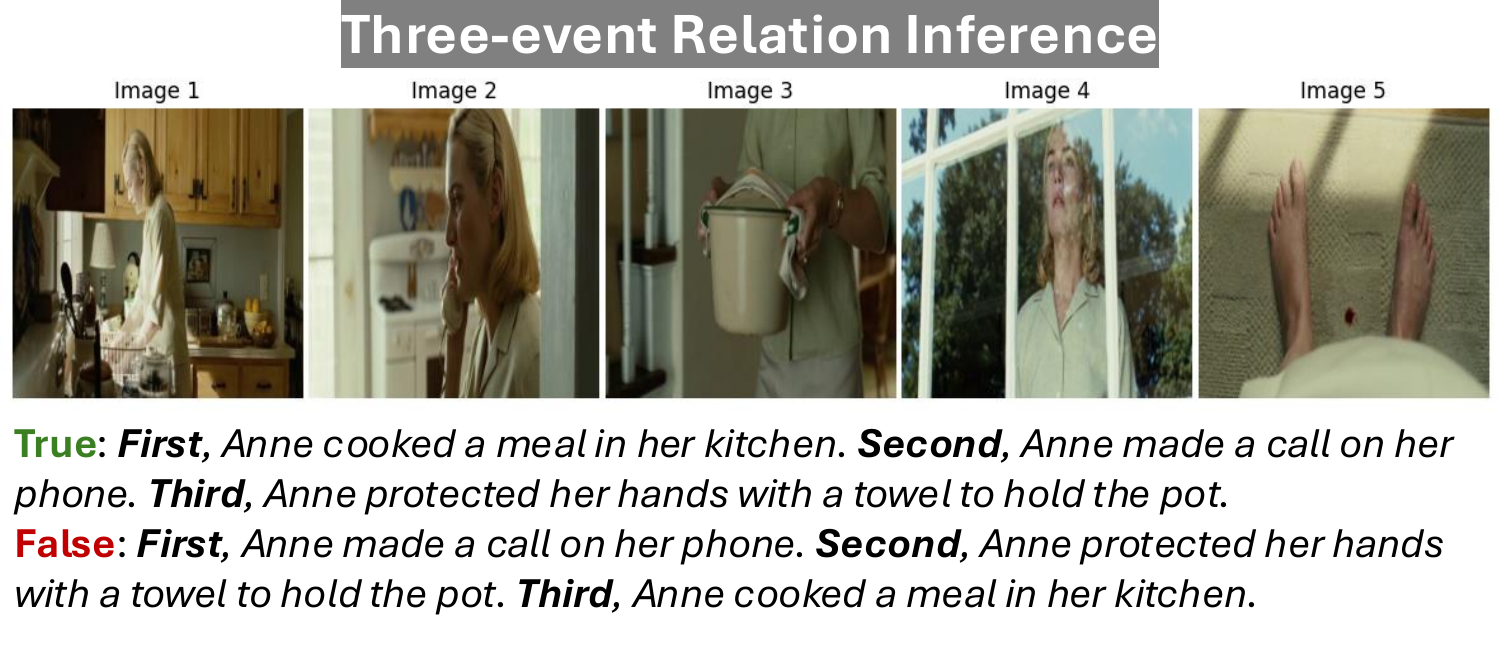}
    \end{subfigure}
    
    \begin{subfigure}{0.495\linewidth}
        \centering
        \includegraphics[width=\linewidth, trim=0 130 0 0, clip]{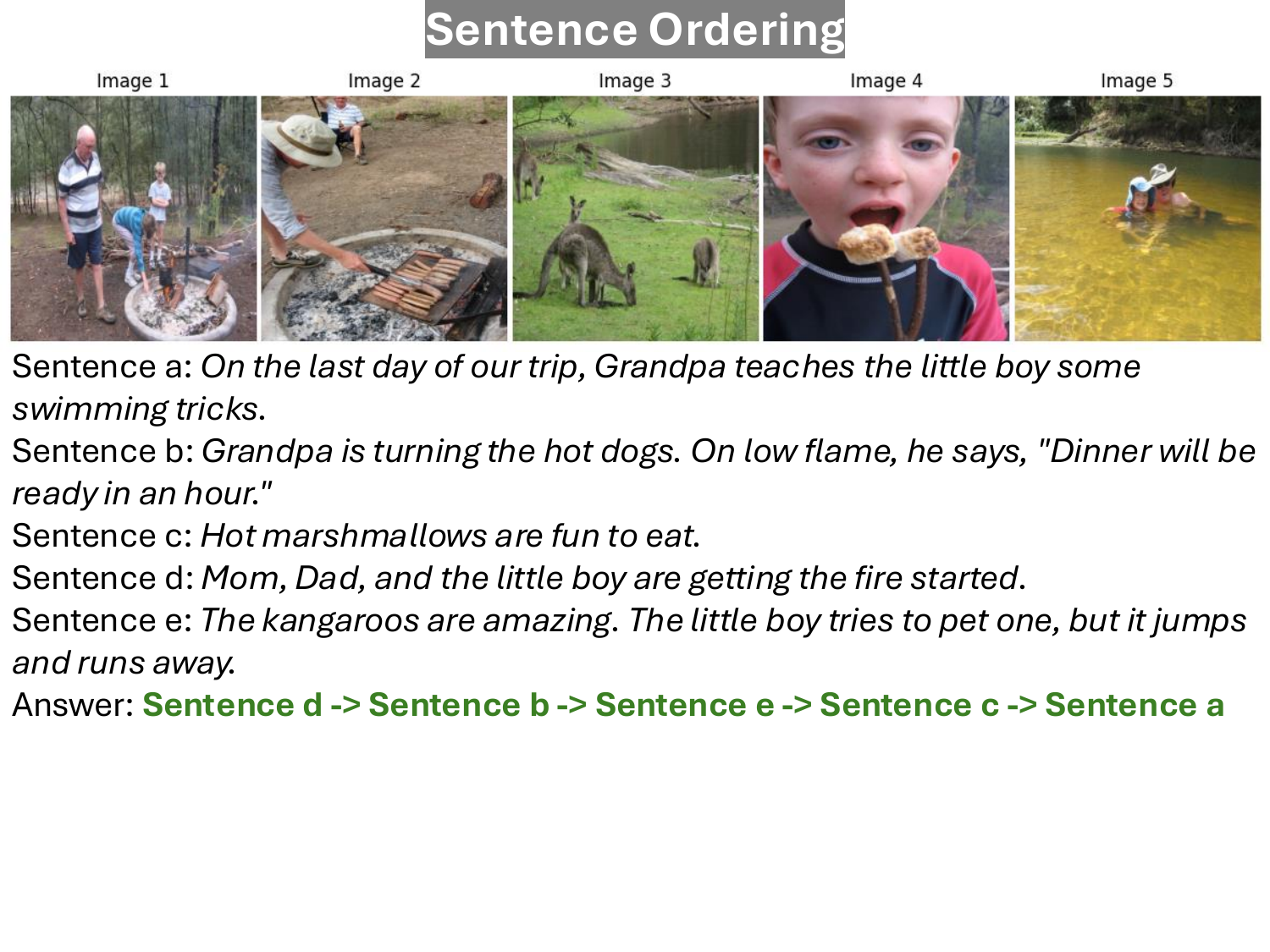}
    \end{subfigure}
    \begin{subfigure}{0.49\linewidth}
        \centering
        \includegraphics[width=\linewidth, trim=0 130 0 0, clip]{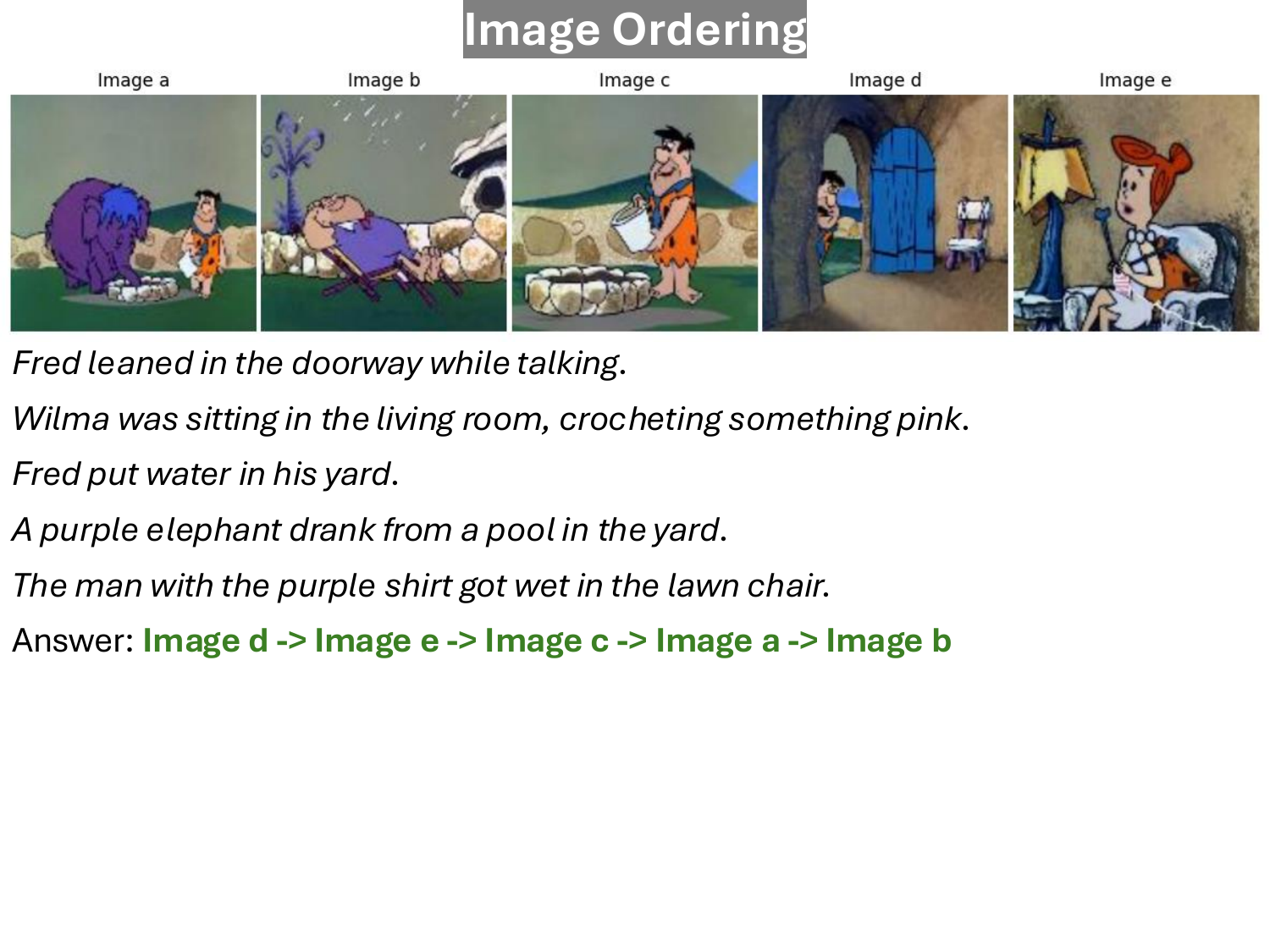}
    \end{subfigure}

    \caption{Illustrative examples from the main tests of TempVS benchmark. Additional examples are provided in \autoref{append:more-examples}.}
    \label{fig:illu-examples}
\end{figure*}

\section{Introduction}
Multimodal Large Language Models (MLLMs) \citep{achiam2023gpt, team2024gemini, liu2024llava} have demonstrated remarkable performance in various vision and language tasks. At the same time, 
the need for standardized evaluation frameworks has become increasingly critical.
Most existing benchmarks focus on settings involving a single image
\citep{mme2023fu, li2024seed, yue2024mmmu, liu2024mmbench, lumathvista}. 
While some benchmarks consider multi-image settings \citep{Jiang2024MANTISIM, fu2024blink, ying2024mmtbench, li2024seed},
they mainly focus on cross-image recognition and reference. To date, relatively little attention has been paid to more complex tasks, such as grounded temporal understanding and reasoning in multiple images.


Some recent studies have assessed MLLMs' temporal comprehension across multiple images, but certain limitations remain. 
First, some tasks can be resolved by relying on a single image rather than a sequence \citep{liu2024mibench, ying2024mmtbench}. 
For example, given an image sequence, determining whether ``pulling a wind-up toy so it continues moving forward or quickly stops'' could be answered only based on the final image.
Second, some tasks depend heavily on commonsense or world knowledge \citep{wang2025muirbench, meng2025mmiu}, such as rearranging a set of shuffled images into the correct sequence of cooking steps.
Third, some benchmarks \citep{dingjie2024milebench} use distractor options absent from the images, allowing the models to infer ground-truth answers based on the presence of objects.
These factors may result in the benchmarks failing to truly assess the model's understanding of temporal sequences.
In addition, none of the existing benchmarks are designed for multi-event scenarios, making them inadequate for evaluating complex temporal sequences and relations in MLLMs.

As a result, a question arises: \textit{Do existing MLLMs really understand time by accurately aligning the order of events in language and image sequences?}
To address this, we propose \textbf{TempVS}, a benchmark for multi-event \textbf{Temp}oral grounding and reasoning in \textbf{V}isual \textbf{S}tory image sequences.
TempVS contains 2,085 image sequences (9,803 images) across cartoon animations, movies and daily-life albums, with 15,192 multiple-choice questions.
TempVS features three temporal understanding and reasoning tasks: event relation inference, sentence ordering, and image ordering (as shown in \autoref{fig:illu-examples}). These tasks are accompanied by grounding tasks to check whether the model can match the exact image consistent with a single text description.
We select image sequences making up visual stories, each containing several events that are temporally related yet relatively independent, in the sense that no event could easily be predicted from preceding events. This makes it hard to resolve the tasks without considering both linguistic and visual modalities. 
TempVS challenges models to reason about event order in the image sequence and text (e.g., a sentence describing two events using \textit{before} or \textit{after}, or a story) and integrate both.

We extensively evaluate 38 MLLMs, including open-source models ranging from 0.5B to 78B (e.g., LLaVA-OneVision, InternVL2.5, Qwen2-VL, Phi-3.5-vision, DeepSeek-vl2, LLaVA-NeXT-Video) and the closed-source GPT-4o. We show that
TempVS is highly challenging for SOTA models,
especially on event relation inference and image ordering tasks.
In particular, while models can accurately ground individual events to images, their performance on the main tasks, which require multimodal reasoning with sequences, remains unsatisfactory. 
We further analyze the impact of the choice of linguistic structure, distance between events, and Chain-of-Thought prompting. 
Our analysis sheds light on future directions for improvement
in architectural design, training objectives, and/or post-training methods to enhance temporal reasoning.

\paragraph{Contributions.} 
(1) We introduce TempVS, a new benchmark for evaluating multi-event temporal grounding and reasoning ability in image sequences for MLLMs.
(2) We extensively evaluate 38 MLLMs from different model families and sizes, highlighting the performance gap compared to human annotators.
(3) Our findings in evaluation results and fine-grained analysis suggest potential pathways for future improvements.

\section{Related Work}

\paragraph{Multimodal Large Language Models}
Progress in large language models (LLMs) \citep{achiam2023gpt, llama3modelcard, touvron2023llama, team2024gemini} has provided impetus to the development of multimodal LLMs which process both visual and textual information.
State-of-the-art MLLMs \citep{achiam2023gpt, DBLP:conf/nips/Dai0LTZW0FH23, team2024gemini, liu2024llava, abdin2024phi, Qwen2VL, chen2024internvl} are built upon LLMs with an integrated visual encoder and a connection module. These models surpass the earlier generation of multimodal models, which were typically based on BERT-type architectures, on many downstream tasks \cite{bugliarello-etal-2023-measuring}.
While some studies have focused on training MLLMs to interpret multiple images using interleaved image-text datasets \citep{Jiang2024MANTISIM, huang2024sparkles, li2024llavaonevisioneasyvisualtask,li2024llavanextinterleavetacklingmultiimagevideo}, their capability to understand and reason about multi-event temporal relationships in sequential visual data remains largely unexplored.

\paragraph{Multi-image Benchmarks}

Multi-image understanding requires MLLMs to compare, analyze and interpret relationships across multiple images~ \citep{li2024survey}.
Benchmarks such as NLVR2 \citep{suhr2019corpus},  BLINK \citep{fu2024blink}, SEED-Bench-2 \citep{li2024seed}, Mantis-Eval \citep{Jiang2024MANTISIM} and MMT-Bench \citep{ying2024mmtbench} cover a subset of multi-image tasks focusing on assessing models' ability to identify similarities and variations across multiple images.
DEMON \citep{li2024finetuning} evaluates the demonstrative instruction-following abilities of MLLMs.
Mementos \citep{DBLP:conf/acl/WangZLLXHYLLBBY24} aims to detect object and behavior hallucinations in descriptive text generated for sequential images.
MileBench \citep{dingjie2024milebench} evaluates MLLMs' performance in long contexts.
MIBench \citep{liu2024mibench} assesses MLLM's ability in multi-image instruction, multimodal knowledge-seeking and in-context learning.
MuirBench \citep{wang2025muirbench} is a comprehensive multi-image understanding benchmark with unanswerable counterparts to test the robustness of MLLMs.
MMIU \citep{meng2025mmiu} incorporates a large number of test questions that cover a diverse array of multi-image tasks and relationships.
We propose TempVS, the first benchmark specifically designed for multi-event temporal understanding and reasoning in image sequences. 
In particular, it is designed to avoid shortcuts such as reliance on single images/frames and commonsense reasoning to bypass full integration of the temporal information in both text and images.

\section{The TempVS Benchmark}
TempVS evaluates models' ability to understand and reason about temporal relations by evaluating the consistency between \textit{textual descriptions of temporally related events} and \textit{the visual event order in an image sequence.}
To achieve this, we create three main tests (MT):
\textbf{event relation inference}, \textbf{sentence ordering}, 
and \textbf{image ordering}.
Additionally, to investigate whether a model’s difficulty arises from challenges in temporal understanding or from more basic grounding issues, we create a corresponding \textbf{grounding test} (GT) (\S\ref{sec:ground_test}) for the main tests.
We present the task curation process and statistics of TempVS benchmark 
in \S\ref{sec:build_benchmark}. 

TempVS is built from existing datasets pairing image sequences with narrative captions. Since captions may vary in their level of detail, we use both the \textbf{original captions} in the source data, and \textbf{simplified captions} in which the main event is extracted from the caption (we describe this process in \S\ref{sec:build_benchmark}). In what follows, we denote an image sequence $S$ consisting of $n$ images, 
their corresponding captions, and extracted events as:
$$\mathcal{S} = [(I_1, C_1, E_1), (I_2, C_2, E_2), \dots, (I_n, C_n, E_n)],$$
where $I_i$ denotes the $i$-th image, $C_i$ its associated original caption, and $E_i$ the extracted event (expressed via the simplified caption).


\begin{table}[h]
\centering
\footnotesize
\resizebox{0.48\textwidth}{!}{
\begin{tabular}{@{}lll@{}}
\toprule
\multicolumn{3}{l}{Statement Template ($i<j<k$ in image sequence)}                                                                                                                                                  \\ \midrule
Two-event   & Pos: $E_j \text{ after } E_i$                                                                     & Neg: $E_i \text{ after } E_j$                                                                     \\ \cmidrule(l){2-3} 
            & Pos: $E_i \text{ before } E_j$                                                                    & Neg: $E_j \text{ before } E_i$                                                                    \\ \cmidrule(l){2-3} 
            & Pos: $E_j. \text{ Earlier, } E_i$                                                                 & Neg: $E_i. \text{ Ealier, } E_j$                                                                  \\ \cmidrule(l){2-3} 
            & Pos: $E_i. \text{ Then, } E_j$                                                                    & Neg: $E_j. \text{ Then, } E_i$                                                                    \\ \cmidrule(l){2-3} 
            & Pos: $E_i. \text{ } E_j.$                                                                         & Neg: $E_j. \text{ } E_i.$                                                                         \\ \midrule
Three-event & \multicolumn{2}{l}{\begin{tabular}[c]{@{}l@{}}Pos: $E_i \text{ before } E_j, \text{and after that, } E_k$\\ Neg: $E_k \text{ before } E_i, \text{and after that, } E_j$\end{tabular}}                 \\ \cmidrule(l){2-3} 
            & \multicolumn{2}{l}{\begin{tabular}[c]{@{}l@{}}Pos: $E_i. \text{ Later, } E_j. \text{Finally, } E_k.$\\ Neg: $E_j. \text{ Later, } E_i. \text{Finally, } E_k.$\end{tabular}}                           \\ \cmidrule(l){2-3} 
            & \multicolumn{2}{l}{\begin{tabular}[c]{@{}l@{}}Pos: $\text{First, }E_i. \text{ Second, } E_j. \text{Third, } E_k.$\\ Neg: $\text{First, }E_i. \text{ Second, } E_k. \text{Third, } E_j.$\end{tabular}} \\ \cmidrule(l){2-3} 
            & Pos: $E_i. \text{ } E_j. \text{ } E_k.$                                                           & Neg: $E_k. \text{ } E_j. \text{ } E_i.$                                                           \\ \bottomrule
\end{tabular}}
\caption{Templates of positive (Pos) and negative (Neg) statements used for \textbf{MT1} event relation inference.}
\label{tab:statement_template}
\end{table}

\subsection{Main Tests and Grounding Test}
\label{sec:main_tests}

\paragraph{MT1: Event Relation Inference} 
MT1 evaluates a model's understanding of the chronological order of events based on an image sequence and a textual description.
The text describes the temporal relation of the events either explicitly through adverbial markers such as \emph{after} or \emph{before}, or implicitly through the surface order of sentences. 
From an image sequence of length $n$, we select either (1) two pairs $\{(I_i, E_i), (I_j, E_j)\}$ where $i<j\leq n$, ensuring that $I_i$ appears before $I_j$ and $E_i$ occurs before $E_j$; or (2) three pairs $\{(I_i, E_i), (I_j, E_j), (I_k, E_k)\}$ where $i<j<k\leq n$, following the same ordering constraints. 
These event-image pairs are not necessarily adjacent in the sequence, resulting in varying distances between them.
We then generate positive and negative statements by applying the templates in \autoref{tab:statement_template} to describe the temporal relations between these events.
\footnote{We only use the textual events to construct the statements in MT1, ignoring the captions.}
The negative statement retains the same event clauses and the temporal conjunction as the positive statement, while swapping the positions of these clauses, such that the text expresses a different temporal order.\footnote{For the negative statements with three events, we randomly select one from the five combinations that is not the same as the positive statement.}
Finally, we derive triples comprising an original image sequence, a positive or negative statement and the corresponding answer (True or False).


\paragraph{MT2: Sentence Ordering}
This task evaluates whether models can correctly reorder a shuffled set of sentences based on the temporal order of events in a given image sequence. Thus, MT2 requires not only an understanding of temporal relations between events but also consideration of the text's coherence and fluency.
Models are tasked to select the correct sentence order from five given options, 
based on an ordered image sequence and a set of shuffled event descriptions. We create versions with both the original captions ($C$) and the extracted events ($E$).

\paragraph{MT3: Image Ordering}
This task mirrors MT2 but focuses on the visual modality, requiring models to rearrange a set of shuffled images into the correct temporal order based on the given textual description.
Similar to MT2, we also examine whether different text styles 
(i.e., original captions or extracted event descriptions) 
could affect the model's ability to determine the correct order.

\paragraph{GT: Grounding Test}
\label{sec:ground_test}
As a prerequisite for solving one of the main tests, we assume that MLLMs should be able to match an event description with the corresponding image in a multi-image sequence. Specifically, given an event description $E_i$ or a caption $C_i$ and image sequence $[I_1, I_2, \dots, I_n]$, models are required to identify the index of the image (i.e., $I_i$) that best corresponds to the given textual description.
The motivation for conducting grounding tests is as follows: If a model passes the grounding tests but fails the corresponding main tests, it indicates that the model struggles with understanding temporal order, even if it can accurately recognize and associate visual and linguistic elements. 
In contrast, if a model performs well on the main tests but fails the corresponding grounding tests, it may suggest that its success stems not from the true temporal grounding or reasoning but rather from leveraging statistical patterns, correlations, or systematic biases learned during training.

\subsection{Benchmark Curation and Statistics}
\label{sec:build_benchmark}

\paragraph{Data Sources}
Given our objective of evaluating whether models understand the chronological order of events across image sequences and language, we require data containing multiple images that form a temporal sequence presenting events.
We choose four visual story datasets: FlintstonesSV \citep{gupta2018imagine}, PororoSV \citep{li2019storygan}, VIST \citep{huang2016visual} and VWP \citep{hong2023visual}.
FlintstonesSV and PororoSV, designed for story visualization, contain annotated frames from cartoon animations.\footnote{For the major characters in FlintstonesSV and PororoSV, we provide descriptions of their appearances to match them with their names.} VIST, built for visual storytelling, originates from Flickr albums with user-uploaded daily-life photos. VWP features movie scene sequences paired with aligned synopses.
This collection with rich styles and diverse domains plays a crucial role in assessing MLLMs' multi-event temporal grounding and reasoning capabilities.

\paragraph{Dataset Filtering}
To ensure that each image sequence contains a sufficient number of character-centered visual events,
we use Detectron2\footnote{\url{https://ai.meta.com/tools/detectron2/}} to detect and retain image sequences where \texttt{PERSON} can be detected in at least 60\% of the images.
To avoid temporal overlap between any two events, we remove any image sequence whose captions contain stative verbs such as \textit{`belong'}, \textit{`love'} and \textit{`exist'}. To minimize ambiguity, we remove captions starting with pronouns and compute the BERTScore~\citep{Zhang2020BERTScore} between captions, omitting sequences with highly similar captions. Similarly, we remove image sequences with highly similar images based on the cosine similarity between their CLIP~\citep{radford2021learning} embeddings. Simple events ($E$) are extracted from the original captions ($C$) using GPT-4~\citep{achiam2023gpt}; any sequences with captions from which no event can be extracted are removed. See Appendix~\ref{append:ev-ext} for details of the extraction process.
To ensure that each image-event pair in $\{(I_i, E_i), (I_j, E_j)\}$ remains distinct, we use CLIP to compute cross-modality similarity between different image-text combinations. Ambiguous pairs are filtered based on a threshold, ensuring that within-pair similarity is significantly higher than cross-pair similarity.\footnote{That is, we guarantee that  $sim(I_i, E_i) > sim(I_i, E_j)$ and $sim(I_i, E_i) > sim(I_j, E_i)$ and $sim(I_j, E_j) > sim(I_i, E_j)$ and $sim(I_j, E_j) > sim(I_j, E_i)$.} A similar process is applied to sets of three image-event pairs  $\{(I_i, E_i), (I_j, E_j), (I_k, E_k)\}$.
In \autoref{append:data-filter}, we provide the details of stative verbs list, the similarity thresholds, prompt used for extracting event from original caption, and statistics of the dataset after each filtering step.

\paragraph{Prompt and Option Generation} 

After filtering the datasets, we create 
the positive and negative statements by concatenating the events with the templates shown in \autoref{tab:statement_template} for MT1.
We sequentially apply each template to its corresponding temporal relation group in the dataset, ensuring an even distribution of statement types. 
Moreover, we use ChatGPT\citep{OpenAI2022ChatGPT} to generate variations of different prompt components for different tasks including task instructions, answer requirements and response formats (see \autoref{append:prompts}), which results in a total of 328 possible prompt variations. By incorporating sufficient diversity in prompts, we mitigate the risk of results being influenced by a specific prompt formulation \citep{DBLP:conf/iclr/Sclar0TS24}.
All tests are formed as multiple-choice questions. In MT1, the options are ``True'' and ``False'' with positions alternated across samples (e.g., A. True; B. False. and A. False; B. True.) to prevent position bias \citep{zheng2024large}. In MT2 and MT3, one correct sequence is presented alongside four randomly shuffled incorrect sequences with options labeled ``A'' to ``E''. The Grounding Test uses image indices as answer choices. To ensure fair evaluation, correct answers are evenly distributed across options throughout the benchmark.

\paragraph{Quality Control} 
To discourage ``blind'' models that leverage language biases, we filter examples in the benchmark that could be easily solved based only on the linguistic modality. 
We apply three unimodal LLMs Phi-3.5-mini-instruct [4B] \citep{abdin2024phi}, Llama-3.1 [8B] \citep{llama3modelcard} and Qwen-2.5-instruct [72B] \citep{qwen2.5}, which are popular LLM backbones in current MLLMs.
In MT1 and MT2, we discard a sample if at least two LLMs can answer the question correctly without visual inputs.
A manual check was performed by the authors to exclude ambiguous images, grammatically incorrect and/or semantically implausible statements, and cases where image sequences did not match the text.

\begin{table}[]
\centering
\footnotesize
\resizebox{0.48\textwidth}{!}{
\begin{tabular}{@{}l|ccccc@{}}
\toprule
              & \begin{tabular}[c]{@{}c@{}}MT1\\ (two)\end{tabular} & \begin{tabular}[c]{@{}c@{}}MT1\\ (three)\end{tabular} & \begin{tabular}[c]{@{}c@{}}MT2\\ (event)\end{tabular} & \begin{tabular}[c]{@{}c@{}}MT2\\ (caption)\end{tabular} & MT3  \\ \midrule
FlintstonesSV & 2,104                                                & 916                                                   & 501                                                   & 485                                                     & 565  \\
PororoSV      & 864                                                 & 172                                                   & 320                                                   & 326                                                     & 395  \\
VWP           & 850                                                 & 208                                                   & 274                                                   & 256                                                     & 316  \\
VIST          & 3,742                                                & 830                                                   & 708                                                   & 551                                                     & 809 \\ \midrule
TempVS        & 7,560                                               & 2,126                                                  & 1,803                                                  & 1,618                                                    & 2,085 \\ \bottomrule
\end{tabular}
}
\caption{TempVS benchmark statistics: In MT1, the number indicates the total statements; in MT2, the number of image sequences and corresponding shuffled sentence sets; in MT3, the number of textual events or captions and their associated shuffled image sets. }
\label{tab:benchmark-stat}
\end{table}

\paragraph{Benchmark Statistics}
\label{sec:benchmark_stat}
TempVS consists of 2,085 distinct image sequences with corresponding original captions and extracted events.
\autoref{tab:benchmark-stat} shows the statistics of each task from each data source in TempVS. 
Most image sequences in the benchmark contain 5 images each, except for 61 sequences from VWP, which have 6 to 9 images.

\section{Experiments}

\subsection{Experimental Setup}

\paragraph{Models} We evaluate a diverse family of state-of-the-art models with various sizes (ranging from 0.5B to 78B) and different vision and LLM backbones. 
We select DeepSeek-vl2 [3B/16B] \citep{wu2024DeepSeekvl2mixtureofexpertsvisionlanguagemodels}, InternVL2.5 [1B/8B/26B/78B] \citep{chen2024internvl}, Janus-Pro [1B/7B]\citep{chen2025januspro}, LLaVA-NeXT-Interleave [0.5B/7B] \citep{li2024llavanextinterleavetacklingmultiimagevideo},
LLaVA-OneVision [0.5B/7B/72B] \citep{li2024llavaonevisioneasyvisualtask},
LLaVA-NeXT-Video [7B/34B] \citep{zhang2024llavanextvideo}, LongVA [7B] \citep{DBLP:journals/corr/abs-2406-16852}, Mantis[8B] \citep{Jiang2024MANTISIM}, Phi-3-vision [4B], Phi-3.5-vision [4B] \citep{abdin2024phi}, and Qwen2-VL [2B/7B/72B] \citep{Qwen2VL}. We also evaluate GPT-4o [2024-11-20]. 
The implementation details are provided in Appendix \ref{append:implem}.\footnote{In the evaluation, we horizontally combine multiple images of a sequence into one image. In our preliminary experiments, we attempted to input the images into the model sequentially, one at a time, and observed little difference in performance compared to merging them into a single input (see results in \autoref{tab:sep-results}). 
In the released TempVS benchmark, both the individual-image inputs and the combined multi-image inputs are provided.
}


\paragraph{Evaluation Metrics}
For multiple-choice questions, we benchmark MLLMs' performance using accuracy as the metric for predicted options. Accuracy scores are reported for both the main tests (MT) and their corresponding grounding tests (GT). In MT1, where each question relates to either two or three events, there is a GT for each separate event. In MT2 and MT3, each question corresponds to a number of grounding tests equal to the number of images in the sequence.
Additionally, we introduce a stricter metric, GT$_{strict}$, which assesses the number of image sequences where a model passes all corresponding grounding tests.\footnote{Appendix \ref{append:full-results} provides overall grounding test results, showing models' ability to precisely locate the correct image within a sequence based on textual input in general.} To examine the relationship between main test and grounding test performance, we further report MT|GT$_{strict}$, where a model's success on a main test is considered valid only if it passes all corresponding grounding tests.

\begin{table*}[]
\centering
\resizebox{\textwidth}{!}{
\begin{tabular}{cccccccccccccccccc}
\hline
\multicolumn{2}{c}{}            & \multicolumn{3}{c}{Two-event Relation (MT1)}        & \multicolumn{3}{c}{Three-event Relation (MT1)}      & \multicolumn{3}{c}{Sentence Ordering - event (MT2)}  & \multicolumn{3}{c}{Sentence Ordering - caption (MT2)} & \multicolumn{2}{c}{Image Ordering - event (MT3)} & \multicolumn{2}{c}{Image Ordering - caption (MT3)} \\
\multicolumn{2}{c}{}            & GT$_s$        & MT            & MT|GT$_s$     & GT$_s$        & MT            & MT|GT$_s$     & GT$_s$        & MT            & MT|GT$_s$      & GT$_s$         & MT             & MT|GT$_s$     & MT                  & MT|GT$_s$            & MT                    & MT|GT$_s$            \\ \hline
\rowcolor[HTML]{9B9B9B} 
Random                &         & 4             & 50            &               & 0.6           & 50            &               & 0.032         & 20            &                & 0.032          & 20             &               & 20                  &                      & 20                    &                      \\
\rowcolor[HTML]{EFEFEF} 
DeepSeek-vl2          & 3B      & 20.8          & 49.7          & 49.7          & 10.5          & 49.8          & 50.6          & 0.8           & 19.5          & 25.0           & 1.5            & 18.2           & 17.9          & 20.4                & 42.9                 & 20.7                  & 12.5                 \\
\rowcolor[HTML]{EFEFEF} 
                      & 16B      & 14.2          & 43.1          & 42.2          & 6.7           & 44.4          & 44.4          & 0.4           & 15.7          & 14.3           & 0.6            & 17.1           & 18.2          & 16.6                & 0.0                  & 15.5                  & 0.0                  \\
InternVL2.5           & 26B     & 46.0          & 57.1          & 58.4          & 39.6          & 58.4          & 60.2          & 10.2          & 56.7          & 73.6           & 13.1           & 63.0           & 76.9          & 26.7                & 27.3                 & 31.7                  & 35.9                 \\
                      & 26B-MPO & 51.3          & \textbf{60.3} & \textbf{62.1} & 46.0          & {\ul 62.1}    & {\ul 64.7}    & 12.6          & {\ul 69.9}    & 90.8           & 17.0           & {\ul 76.9}     & {\ul 87.3}    & {\ul 34.4}          & 39.7                 & 39.5                  & 43.7                 \\
                      & 78B     & 51.6          & 54.2          & 55.3          & 47.3          & 56.8          & 57.0          & 13.6          & 67.0          & 84.9           & 18.5           & 71.1           & 83.9          & 31.1                & 40.0                 & 38.5                  & 47.1                 \\
                      & 78B-MPO & {\ul 58.8}    & 58.5          & 59.9          & {\ul 56.5}    & 61.4          & 62.6          & {\ul 18.4}    & \textbf{79.8} & {\ul 96.6}     & {\ul 25.9}     & \textbf{86.3}  & \textbf{96.4} & \textbf{41.0}       & {\ul 48.8}           & \textbf{53.8}         & \textbf{69.7}        \\
\rowcolor[HTML]{EFEFEF} 
Janus-Pro             & 1B      & 2.7           & 48.3          & 48.1          & 0.7           & 46.5          & 42.6          & 0.0           & 18.6          & -              & 0.1            & 19.8           & 0.0           & 22.5                & 0.1                    & 22.3                  & 0.0                  \\
\rowcolor[HTML]{EFEFEF} 
                      & 7B      & 4.3           & 35.1          & 34.1          & 0.4           & 32.9          & 39.3          & 0.0           & 17.1          & -              & 0.0            & 15.3           & -             & 20.9                & -                    & 21.0                  & -                    \\
LLaVA-NeXT-Interleave & 0.5B    & 2.5           & 49.8          & 47.8          & 0.2           & 50.4          & 50.0          & 0.0           & 20.7          & -              & 0.0            & 20.7           & -             & 20.2                & -                    & 20.8                  & -                    \\
                      & 7B      & 13.0          & 51.6          & 52.4          & 7.2           & 50.1          & 49.8          & 0.3           & 25.1          & 16.7           & 0.4            & 27.0           & 0.0           & 20.9                & 16.7                 & 20.0                  & 22.2                 \\
\rowcolor[HTML]{EFEFEF} 
LLaVA-OneVision-ov    & 0.5B    & 8.6           & 45.3          & 45.5          & 2.9           & 48.1          & 47.6          & 0.1           & 18.8          & 0.0            & 0.1            & 18.4           & 0.0           & 19.4                & \textbf{100.0}       & 19.0                  & 0.0                  \\
\rowcolor[HTML]{EFEFEF} 
                      & 7B      & 32.8          & 56.0          & 58.0          & 26.0          & 57.5          & 59.8          & 4.5           & 44.2          & 41.2           & 6.8            & 46.9           & 47.6          & 21.3                & 14.3                 & 21.6                  & 19.8                 \\
\rowcolor[HTML]{EFEFEF} 
                      & 72B     & 46.4          & 59.3          & \textbf{62.1} & 40.5          & 61.5          & 63.5          & 9.8           & 65.2          & 81.8           & 14.0           & 75.1           & 86.6          & 27.6                & 31.8                 & 29.1                  & 36.5                 \\
LLaVA-NeXT-Video      & 7B      & 5.8           & 46.0          & 46.0          & 1.4           & 44.9          & 47.0          & 0.0           & 19.0          & -              & 0.0            & 18.2           & -             & 21.0                & -                    & 21.3                  & -                    \\
                      & 34B     & 6.5           & 58.5          & 58.4          & 1.6           & 59.5          & 57.6          & 0.1           & 31.8          & \textbf{100.0} & 0.0            & 33.4           & -             & 19.8                & -                  & 20.0                  & -                    \\
\rowcolor[HTML]{EFEFEF} 
LongVA                & 7B      & 8.7           & 54.7          & 56.1          & 2.1           & 56.0          & 61.7          & 0.2           & 34.2          & 66.7           & 0.2            & 35.3           & 50.0          & 19.5                & 0.1                    & 19.0                  & -                    \\
Mantis-Idefics        & 8B      & 12.2          & 51.9          & 53.3          & 4.1           & 52.0          & 51.6          & 0.1           & 22.2          & 0.0            & 0.2            & 20.8           & 0.0           & 18.6                & 0.0                  & 19.2                  & {\ul 50.0}           \\
\rowcolor[HTML]{EFEFEF} 
Phi-3.5-vision               & 3.4B    & 4.0           & 49.0          & 47.7          & 0.8           & 48.8          & 48.3          & 0.0           & 23.1          & -              & 0.0            & 25.4           & -             & 19.2                & -                    & 18.3                  & -                    \\
Qwen2-VL-Instruct     & 7B      & 32.4          & 54.0          & 55.4          & 21.2          & 53.6          & 55.3          & 3.4           & 42.5          & 64.6           & 4.4            & 44.6           & 61.3          & 23.1                & 20.4                 & 24.6                  & 35.9                 \\
                      & 72B     & 31.7          & 54.0          & 56.4          & 20.6          & 55.6          & 60.6          & 3.7           & 46.3          & 64.3           & 5.0            & 55.1           & 70.0          & 26.5                & 47.3                 & 28.1                  & 47.4                 \\
\rowcolor[HTML]{EFEFEF} 
GPT-4o                & API     & \textbf{60.3} & {\ul 58.3}    & {\ul 60.1}    & \textbf{57.0} & \textbf{64.5} & \textbf{66.4} & \textbf{18.6} & 53.4          & 53.9           & \textbf{28.6}  & 61.5           & 55.3          & 22.6                & 23.5                 & 23.0                  & 23.5                 \\ \hline
                      &         &               &               &               &               &               &               &               &               &                &                &                &               &                     &                      &                       &                     
\end{tabular} 
}
\caption{Zero-shot average accuracy performance of 11 popular MLLMs families with 21 variants on TempVS benchmark on strict grounding test (GT$_s$), main test (MT) and the main test when all corresponding grounding tests pass (MT|GT$_s$). Best models per metric are marked in boldface and the second best models are underlined. }
\label{tab:main-results}
\end{table*}

\subsection{Main Results}
Results for 21 state-of-the-art  MLLMs are shown in \autoref{tab:main-results}. 
For the full quantitative results of all tested 38 MLLMs, we refer to Appendix \ref{append:full-results}. 
Our main observations of the empirical results are as follows:

\paragraph{TempVS is challenging even for SOTA MLLMs}
InternVL2.5-26B-MPO achieves the highest performance on the two-event relation inference, while GPT-4o leads in the three-event relation inference. InternVL2.5-78B-MPO outperforms other models in both sentence ordering and image ordering tasks.
However, most models with parameters less than or equal to 7B exhibit random chance accuracies of approximately  50\% (for MT1) and 20\% (for MT2 and MT3).
Most MLLMs perform similarly on two-event and three-event relation inference tasks, while for the strongest models (such as InternVL2.5[26B/78B], LLaVA-OneVision-ov-72B and GPT-4o), their performance is even slightly better on the understanding of three-event statements.
In addition, sentence ordering is a relatively simple task (with the highest accuracy at 86.3\%), while image ordering is a significantly more challenging task (with the highest accuracy only at 53.8\%).
GPT-4o performs substantially worse than several of the best-performing open-source models in both sentence and image ordering tasks.
In terms of language type, we find that sentence and image ordering are easier with original captions than with extracted events. This may indicate that models might leverage additional contextual details and temporal cues from the original captions, which are unavailable in the simpler extracted event descriptions.

\begin{figure*}[h] 
    \centering
    \includegraphics[width=\textwidth]{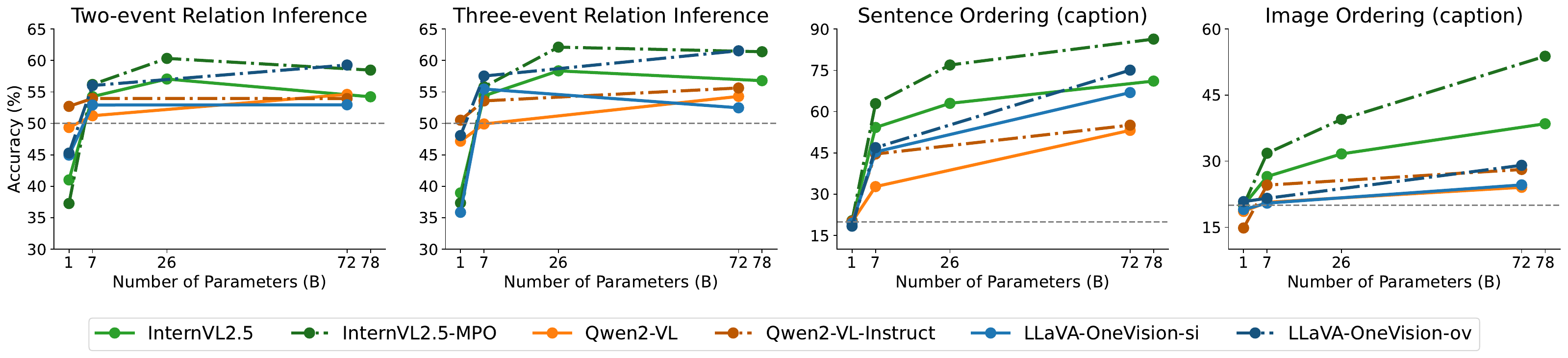} 
    \caption{Illustrative TempSV benchmark results of selected models with different number of parameters.}
    \label{fig:larger_models} 
\end{figure*}

\paragraph{Both scale and post-training help}
As shown in \autoref{fig:larger_models}, accuracy generally improves with model size. 
However, the marginal gains diminish as models get larger.
In the two-event and three-event relation inference, this effect is more evident,
with smaller models (7B or 26B) already attaining competitive and in some cases higher accuracy compared to larger models ( $>$ 70B).
Additionally, the two ordering tasks exhibit larger performance gaps between smaller and larger models, which can be attributed to the superior long-range reasoning capabilities of the more powerful LLM backbones in larger MLLMs.
Surprisingly, DeepSeek-VL2 [3B/16B] and Janus-Pro [1B/7B] are exceptions, as their smaller models outperform the larger ones in most cases.

Our results also highlight the importance of high-quality post-training: 
under the same model sizes, InternVL2.5-MPO consistently outperforms InternVL2.5 across all evaluated tasks, especially when the model parameters exceed 7B. These results indicate that Mixed Preference Optimization (MPO) \citep{chen2024internvl} effectively enhances the overall multimodal temporal understanding and reasoning capabilities.
Similarly, models fine-tuned using Direct Preference Optimization (DPO) are consistently better than their SFT-only counterparts. Other results comparing LLaVA-NeXT-Video and LongVA families are provided in Appendix \ref{append:full-results}.
We also see a positive impact of instruction tuning on complex reasoning tasks such as image ordering (e.g., Qwen2-VL-Instruct outperforms Qwen2-VL across all tasks).

\paragraph{Event grounding does not guarantee understanding of temporal relations}
In MT1, we observe a slight difference between MT and MT|GT$_{strict}$.
This suggests that even if a model can accurately identify the image corresponding to every single event within a sequence, it may still lack the ability to understand the chronological order of events in text. In short, grounding textual descriptions and reasoning about their temporal relations require different capabilities.
Among all grounding tests, GPT-4o performs the best, but it lags significantly behind top-tier open-source models like InternVL2.5-MPO[26B/78B] especially in the two ordering tasks (MT2 and MT3).
In MT2 and MT3, MT|GT$_{strict}$ accuracy improves substantially over MT for most large MLLMs, 
confirming the fundamental role of visual grounding in sentence or image ordering tasks.
For instance, the MT|GT$_{strict}$ accuracy scores of InternVL2.5-78B-MPO are higher than MT by 16.8\% (event) and 10.1\% (caption) in MT2, and by 7.8\% (event) and 15.9\% (caption) in MT3.
Meanwhile, for small models like DeepSeek-vl2[3B], MT|GT$_{strict}$ accuracy is sometimes even smaller than MT accuracy. 
This indicates some dependency of these models' reasoning abilities on their grounding capability, though this is unlikely to be the only factor affecting performance.

\begin{table}[]
\centering
\resizebox{0.49\textwidth}{!}{
\begin{tabular}{@{}lccc@{}}
\toprule
                           & \multicolumn{1}{l}{\# Image seqs} & \multicolumn{1}{l}{Accuracy} & \multicolumn{1}{l}{Fleiss' kappa} \\ \midrule
Two-event Relation      & 60                       & 82.5                         & 0.728                              \\
Three-event Relation    & 60                       & 81.6                         & 0.689                              \\
Sentence Ordering (event) & 40                       & 81.2                         & 0.751                              \\
Sentence Ordering (caption) & 40                       & 89.3                         & 0.764                             \\
Image Ordering (event)    & 40                       & 79.1                         & 0.827                              \\
Image Ordering (caption)     & 40                       & 77.9                        & 0.742                              \\ \bottomrule
\end{tabular}}
\caption{Results of human evaluation on all main tests.
}
\label{tab:human-results}
\end{table}

\paragraph{Gap with human performance}
To evaluate the quality and estimate the difficulty of TempVS, we perform a human assessment on 280 randomly selected image sequences, covering all fine-grained statement types of MT1 as well as both language types ($C$ and $E$) for MT2 and MT3.
We recruited 36 annotators on the Prolific platform and collected three responses for each image sequence, yielding 840 responses collected in total. Further details are provided in \autoref{append:human-survey}. 

There is a large gap between human average performance (\autoref{tab:human-results}) and SOTA MLLMs. 
For the tasks of two-event and three-event relation inference and image ordering, there remains plenty of room for improvement.
However, for the sentence ordering task, the strongest model InternVL2.5-78B-MPO is already close to human performance.
Humans exhibit substantial agreement among themselves on the main tasks, with Fleiss' kappa \citep{landis1977measurement} above 0.68 across the board.

\begin{table*}[]
\resizebox{\textwidth}{!}{
\begin{tabular}{@{}ccccccccccccccc@{}}
\toprule
                                                                                     & Models           & \multicolumn{2}{c}{InternVL2.5-26B-MPO}                     & \multicolumn{2}{c}{InternVL2.5-78B-MPO}                     & \multicolumn{2}{c}{llava-onevision-72b-ov} & \multicolumn{2}{c}{LLaVA-NeXT-Video-34B}                     & \multicolumn{2}{c}{GPT-4o}                                  & \multicolumn{3}{c}{Top-5 Average}                                  \\
\multirow{-2}{*}{Tasks}                                                              & Statement Type   & Pos                          & Neg                          & Pos                          & Neg                          & Pos                              & Neg     & Pos                           & Neg                          & Pos                          & Neg                          & Pos                          & Neg                          & Both \\ \midrule
                                                                                     & after            & 59.9                         & \cellcolor[HTML]{C0C0C0}62.9 & \cellcolor[HTML]{C0C0C0}70.7 & 49.1                         & \cellcolor[HTML]{C0C0C0}71.4     & 50.6    & \cellcolor[HTML]{C0C0C0}77.9  & 37.1                         & 56.2                         & \cellcolor[HTML]{C0C0C0}60.4 & \cellcolor[HTML]{C0C0C0}67.2 & 52.0                         & 59.6 \\
                                                                                     & before           & 66.4                         & \cellcolor[HTML]{C0C0C0}66.7 & \cellcolor[HTML]{C0C0C0}71.6 & 57.9                         & \cellcolor[HTML]{C0C0C0}70.6     & 59.0    & \cellcolor[HTML]{C0C0C0}77.1  & 41.4                         & 53.3                         & \cellcolor[HTML]{C0C0C0}73.4 & \cellcolor[HTML]{C0C0C0}67.8 & 59.7                         & 63.8 \\
                                                                                     & earlier          & \cellcolor[HTML]{C0C0C0}70.8 & 47.0                         & \cellcolor[HTML]{C0C0C0}74.3 & 36.8                         & \cellcolor[HTML]{C0C0C0}74.1     & 41.5    & \cellcolor[HTML]{C0C0C0}85.9  & 28.8                         & \cellcolor[HTML]{C0C0C0}69.1 & 38.5                         & \cellcolor[HTML]{C0C0C0}74.8 & 38.5                         & 56.7 \\
                                                                                     & then             & \cellcolor[HTML]{C0C0C0}65.5 & 59.3                         & \cellcolor[HTML]{C0C0C0}73.3 & 44.8                         & \cellcolor[HTML]{C0C0C0}71.3     & 50.0    & \cellcolor[HTML]{C0C0C0}86.4  & 35.2                         & 61.0                         & \cellcolor[HTML]{C0C0C0}63.1 & \cellcolor[HTML]{C0C0C0}71.5 & 50.5                         & 61.0 \\
\multirow{-5}{*}{\begin{tabular}[c]{@{}c@{}}Two-event \\ Relation\end{tabular}}   & implict      & \cellcolor[HTML]{C0C0C0}65.3 & 38.7                         & \cellcolor[HTML]{C0C0C0}74.4 & 31.6                         & \cellcolor[HTML]{C0C0C0}74.4     & 30.1    & \cellcolor[HTML]{C0C0C0}86.4  & 28.3                         & \cellcolor[HTML]{C0C0C0}64.0 & 43.4                         & \cellcolor[HTML]{C0C0C0}72.9 & 34.4                         & 53.6 \\ \midrule
                                                                                     & before/after      & 66.7                         & \cellcolor[HTML]{C0C0C0}72.3 & \cellcolor[HTML]{C0C0C0}70.7 & 64.0                         & \cellcolor[HTML]{C0C0C0}77.7     & 54.6    & 71.2                          & \cellcolor[HTML]{C0C0C0}72.0 & 55.7                         & \cellcolor[HTML]{C0C0C0}78.2 & 65.1                         & \cellcolor[HTML]{C0C0C0}67.2 & 66.2 \\
                                                                                     & first/second/third & 55.2                         & \cellcolor[HTML]{C0C0C0}70.4 & 65.4                         & \cellcolor[HTML]{C0C0C0}67.5 & \cellcolor[HTML]{C0C0C0}76.5     & 45.9    & \cellcolor[HTML]{C0C0C0}68.4  & 64.1                         & 50.4                         & \cellcolor[HTML]{C0C0C0}78.9 & 60.4                         & \cellcolor[HTML]{C0C0C0}65.5 & 63.0 \\
                                                                                     & later/finally     & \cellcolor[HTML]{C0C0C0}75.2 & 56.3                         & \cellcolor[HTML]{C0C0C0}87.5 & 32.5                         & \cellcolor[HTML]{C0C0C0}77.4     & 52.8    & \cellcolor[HTML]{C0C0C0}100.0 & 0.1                          & 69.4                         & \cellcolor[HTML]{C0C0C0}68.3 & \cellcolor[HTML]{C0C0C0}79.8 & 43.1                         & 61.5 \\
\multirow{-4}{*}{\begin{tabular}[c]{@{}c@{}}Three-event \\ Relation\end{tabular}} & implict    & \cellcolor[HTML]{C0C0C0}73.3 & 27.1                         & \cellcolor[HTML]{C0C0C0}82.6 & 20.3                         & \cellcolor[HTML]{C0C0C0}78.7     & 28.4    & \cellcolor[HTML]{C0C0C0}100.0 & 0.3                          & \cellcolor[HTML]{C0C0C0}70.3 & 43.7                         & \cellcolor[HTML]{C0C0C0}76.3 & 28.4                         & 52.4 \\ \bottomrule
\end{tabular}
}
\caption{Fine-grained accuracy of different statement types in the two-event and three-event relation inference tasks. ``Pos'' denotes positive examples, while ``Neg'' represents negative examples. The higher accuracy score is highlighted in gray for each pair of positive and negative statements.}
\label{tab:statement_type}
\end{table*}

\subsection{Further Analysis}

\paragraph{Impact of temporal expressions}
We further analyze how the models understand and reason about different types of statements in the two-event and three-event relation inference tasks (MT1).
We select the top five models on these two tasks for comparison (\autoref{tab:statement_type}).
When comparing explicit and implicit temporal event statements (cf. \autoref{tab:statement_template}), we observe that models consistently perform better on the former.
This could be because the presence of temporal adverbs or conjunctions in a sentence helps clarify the order in which events happen. 
The top five models always achieve higher accuracy on positive examples than on negative ones. Despite the even distribution of ``True'' and ``False'' across options ``A'' and ``B'' in our benchmark, models exhibit a tendency to predict ``True'' more frequently. By incorporating adversarial samples, our TempVS benchmark effectively reveals MLLMs' biases toward certain answers, providing a robust assessment of their compositional temporal reasoning capabilities.

We also observe the better performance of models on explicitly marked temporal relations involving {\em before} (resp. {\em after}) and on {\em then} (resp. {\em earlier}). The key difference between these pairs of complementary temporal adverbials is that with {\em before} and {\em then}, the order of events in text mirrors their order in the image sequence, whereas this is not the case with {\em after}/{\em earlier}. We therefore see some evidence of an {\em iconicity effect}: temporal relations are easier for models when the surface order of events mirrors their actual order (in the visual modality). This echoes similar findings in the discourse processing and psycholinguistic literature on narrative comprehension~\cite{Zwaan1998,smith_modes_2003}. It also points to an important avenue for future research in fine-grained multimodal benchmarking, namely, in cases where surface characteristics in two modalities are not perfectly aligned.


\begin{figure}[] 
    \centering
\includegraphics[width=0.48\textwidth]{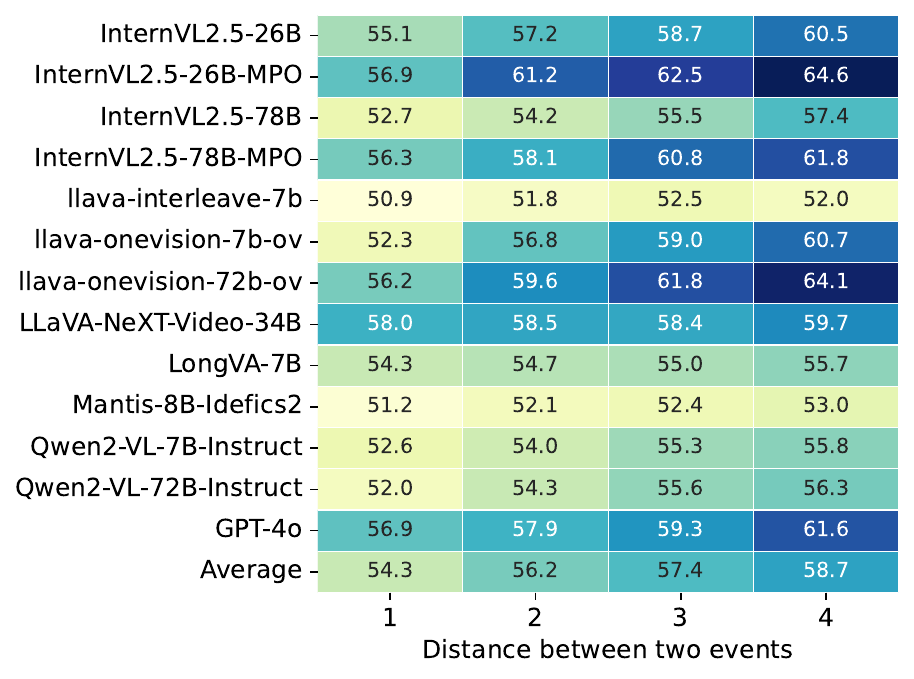} 
    \caption{Accuracy of the two-event relation inference task with different distances between the two events in a sequence.}
    \label{fig:distance} 
\end{figure}

\paragraph{Distance between two events}
\autoref{fig:distance} presents the accuracy of the two-event relation inference task as the distance between events in the original sequence increases from one to four. Nearly all models improve in performance as the distance increases. On the one hand, more distant visual events in the dataset are typically more visually distinct. On the other hand, the models may fail to effectively separate two closely spaced images, even though separators were inserted between images in the input.
\begin{table}[]
\resizebox{0.49\textwidth}{!}{
\begin{tabular}{c|ccc|ccc}
\hline
                     & \multicolumn{3}{c|}{InternVL2.5-78B}                       & \multicolumn{3}{c}{GPT-4o}                                 \\
\multirow{-2}{*}{}   & w/o. CoT & w. CoT & $\Delta$                 & w/o. CoT & w. CoT & $\Delta$                  \\ \hline
Two-event relation (MT1)   & 54.23    & 56.13  & {\color[HTML]{036400} \textbf{+1.90}}  & 58.27    & 60.43  & {\color[HTML]{036400} \textbf{+2.16}}  \\
Three-event relation (MT1) & 56.79    & 57.20  & {\color[HTML]{036400} \textbf{+0.41}}  & 64.52    & 63.38  & {\color[HTML]{CB0000} \textbf{-1.14}}  \\
Sentence Ordering (MT2)   & 71.09    & 85.42  & {\color[HTML]{036400} \textbf{+14.33}} & 61.50    & 81.41  & {\color[HTML]{036400} \textbf{+19.91}} \\
Image Ordering (MT3)      & 38.49    & 47.54  & {\color[HTML]{036400} \textbf{+9.05}}  & 22.97    & 33.77  & {\color[HTML]{036400} \textbf{+10.80}} \\ \hline
\end{tabular}}
\caption{Model performance comparison with and without Chain of Thought (CoT).}
\label{tab:cot}
\end{table}

\paragraph{Prompting with Chain-of-Thought}

Chain-of-Thought (CoT, \citealp{wei2022chain}) is a widely used approach to enhance models' reasoning ability, 
by allowing models to generate intermediate reasoning steps before producing the final answer. It may thus also enhance models' temporal reasoning skills on TempVS.
We conduct CoT experiments using InternVL2.5-78B and GPT-4o.
The detailed prompts are listed in Appendix \ref{append:cot-prompt}.
As shown in \autoref{tab:cot}, CoT yields large gains on sentence and image ordering, but limited improvement for event relation inference (MT1).
This indicates the potential of step-by-step reasoning for ordering tasks. However, simple CoT does not help event relation inference. 
We leave the investigation of methods to enhance models' understanding of this complex task and improve its temporal reasoning capabilities for future work.

\paragraph{Qualitative Case Analysis}

\begin{figure}[ht]
    \centering
    \begin{subfigure}{0.5\textwidth}
        \centering
    \includegraphics[width=\linewidth]{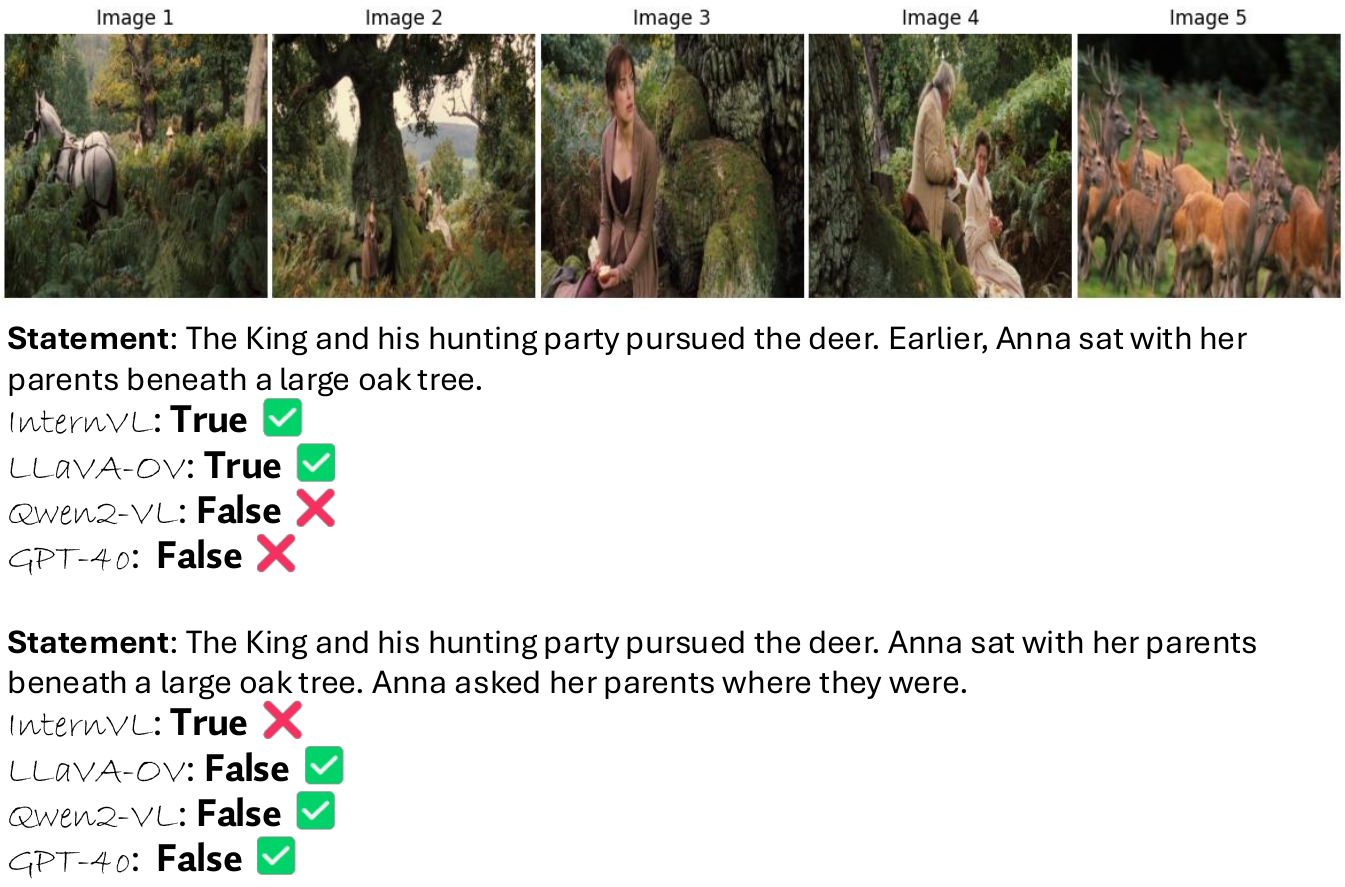}
    \end{subfigure}
    \vspace{0.3em}
    \begin{subfigure}{0.5\textwidth}
        \centering
       \includegraphics[width=\linewidth]{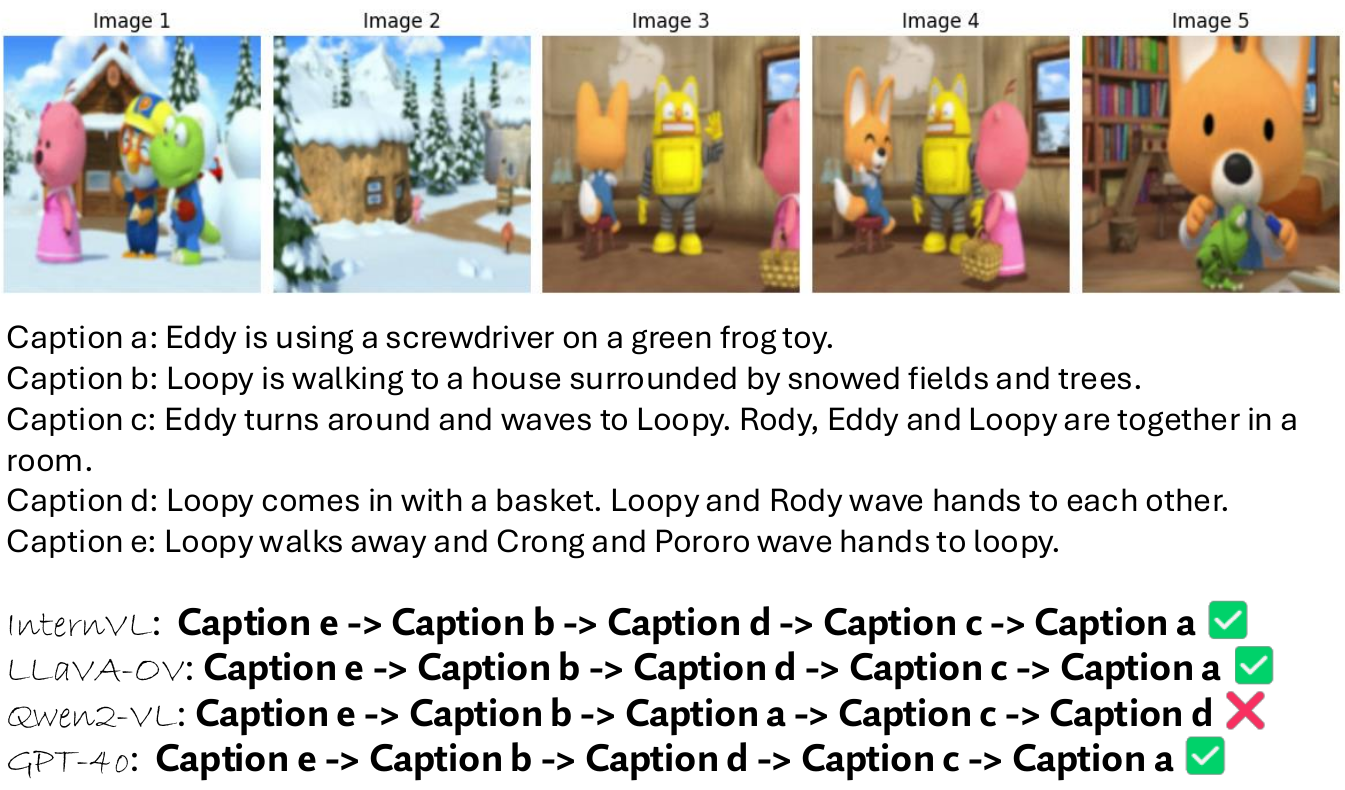}
    \end{subfigure}

    \vspace{0.3em}

    \begin{subfigure}{0.5\textwidth}
        \centering
       \includegraphics[width=\linewidth]{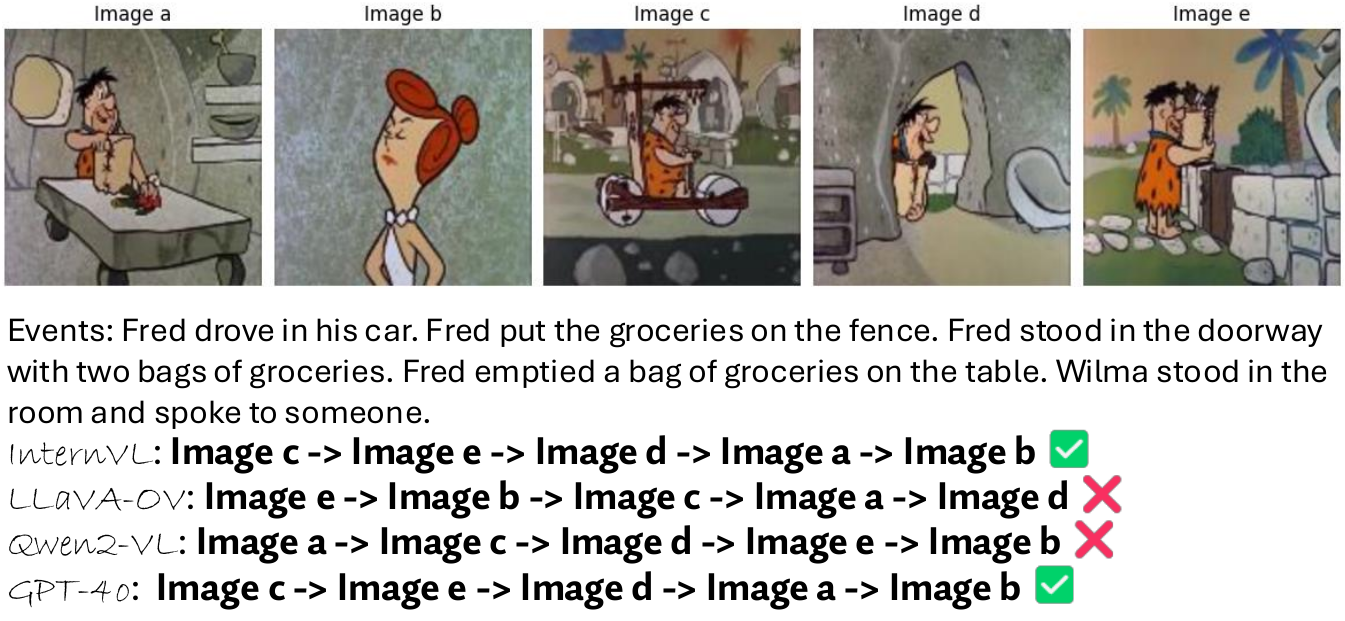}
    \end{subfigure}

    \caption{Qualitative cases of four representative MLLMs on TempVS.}
    \label{fig:quali_cases}
\end{figure}

\autoref{fig:quali_cases} presents task-specific qualitative results among four representative open-source
MLLMs, i.e., InternVL2.5-78B-MPO, LLaVA-OneVision-ov-72b, Qwen2-VL-Instruct-72B and GPT-4o. Further analysis reveals that the unsatisfactory performance of current models stems from: (1) weak event grounding, where models fail to identify the correct image corresponding to an event; (2) poor instruction-following abilities, resulting in outputs that do not follow the required format; (3) a limited understanding of the full sequence, leading to selections that are only partially ordered correctly; and (4) difficulty distinguishing between subtly different images when such  differentiation is necessary for temporal reasoning.

\section{Conclusion}
In this paper, we present a novel and challenging benchmark TempVS, designed to assess multimodal, multi-event temporal reasoning abilities of MLLMs in image sequences.
After evaluating 38 advanced MLLMs, we find that current models typically struggle with reasoning about temporal relations and rearranging shuffled images to the correct order based on a narrative.
Further analysis of linguistic structures, event distance, and Chain-of-Thought reasoning has shed light on promising avenues for future work.
Our study contributes to MLLM development by uncovering weaknesses in multi-event temporal reasoning in multi-image scenarios, while TempVS provides a valuable resource for further research.

\paragraph{Limitations}

Despite a careful examination of publicly available repositories, technical reports and papers, we find no evidence that the evaluated MLLMs were trained on the data included in TempVS. However, for models (such as GPT-4o) that have not fully disclosed or explicitly stated their training data, the possibility of data leakage and contamination remains unclear. This could potentially lead to an overestimation of their advantages.
Additionally, we focus on multiple-choice questions to ensure structured evaluation and clear correctness criteria, following existing multimodal benchmarks \citep{wang2025muirbench, meng2025mmiu, mme2023fu, liu2024mmbench, li2024seed}. 
However, other question types, such as open-ended questions, particularly the evaluation of open-ended generation in multi-image temporal understanding and reasoning, are also worth exploring. 
We leave these for future work. 

\paragraph{Ethical Considerations}
In this study, we employ published datasets and pretrained multimodal large language models, with no known significant ethical concerns regarding their usage. However, we acknowledge that biases in the original image-caption data may influence both the models and their evaluations. Our research has received approval from our institution’s Ethics Board, ensuring compliance with ethical guidelines for human annotation process. Additionally, all collected human-annotated data has been deidentified to protect participants' data privacy and security.

\paragraph{Acknowledgments}
We thank the anonymous reviewers for their valuable comments and constructive feedback. We are also grateful to Letitia Parcalabescu for her insightful suggestions and engaging discussions, which helped us improve the quality of this work. Our thanks go to the members of UU NLP Group for their assistance with the pilot human evaluation experiments and for their helpful suggestions on earlier drafts of the manuscript.



\bibliography{latex/custom}

\clearpage
\appendix

\section{Details in Data Filtering}
\label{append:data-filter}
\subsection{Stative Verbs}
We filter out the samples where any form of a stative verb exists (as displayed in \autoref{tab:verb_forms}) in the captions to avoid describing a state. In contrast, we only keep the examples with dynamic verbs to describe actions.

\begin{table}[ht]
    \centering
\resizebox{0.48\textwidth}{!}{
    \begin{tabular}{l|ccccc}
        \hline
        \textbf{Base Form} & \textbf{Present Participle} & \textbf{3rd Person Singular} & \textbf{Past Tense} & \textbf{Past Participle} \\
        \hline
       wish & wishing & wishes & wished & wished \\
equal & equaling & equals & equaled & equaled \\
signify & signifying & signifies & signified & signified \\
feel & feeling & feels & felt & felt \\
involve & involving & involves & involved & involved \\
sense & sensing & senses & sensed & sensed \\
sound & sounding & sounds & sounded & sounded \\
detest & detesting & detests & detested & detested \\
want & wanting & wants & wanted & wanted \\
see & seeing & sees & saw & seen \\
forget & forgetting & forgets & forgot & forgot \\
matter & mattering & matters & mattered & mattered \\
contain & containing & contains & contained & contained \\
own & owning & owns & owned & owned \\
taste & tasting & tastes & tasted & tasted \\
dislike & disliking & dislikes & disliked & disliked \\
remember & remembering & remembers & remembered & remembered \\
suppose & supposing & supposes & supposed & supposed \\
resemble & resembling & resembles & resembled & resembled \\
think & thinking & thinks & thought & thought \\
envy & envying & envies & envied & envied \\
depend & depending & depends & depended & depended \\
hate & hating & hates & hated & hated \\
know & knowing & knows & knew & known \\
require & requiring & requires & required & required \\
love & loving & loves & loved & loved \\
appreciate & appreciating & appreciates & appreciated & appreciated \\
need & needing & needs & needed & needed \\
concern & concerning & concerns & concerned & concerned \\
span & spanning & spans & spanned & spanned \\
appear & appearing & appears & appeared & appeared \\
owe & owing & owes & owed & owed \\
weigh & weighing & weighs & weighed & weighed \\
disagree & disagreeing & disagrees & disagreed & disagreed \\
become & becoming & becomes & became & become \\
fear & fearing & fears & feared & feared \\
measure & measuring & measures & measured & measured \\
possess & possessing & possesses & possessed & possessed \\
like & liking & likes & liked & liked \\
look & looking & looks & looked & looked \\
imagine & imagining & imagines & imagined & imagined \\
mind & minding & minds & minded & minded \\
belong & belonging & belongs & belonged & belonged \\
loathe & loathing & loathes & loathed & loathed \\
lack & lacking & lacks & lacked & lacked \\
deserve & deserving & deserves & deserved & deserved \\
mean & meaning & means & meant & meant \\
promise & promising & promises & promised & promised \\
believe & believing & believes & believed & believed \\
prefer & preferring & prefers & preferred & preferred \\
cost & costing & costs & costed & costed \\
hope & hoping & hopes & hoped & hoped \\
recognize & recognizing & recognizes & recognized & recognized \\
include & including & includes & included & included \\
support & supporting & supports & supported & supported \\
understand & understanding & understands & understood & understood \\
comprise & comprising & comprises & comprised & comprised \\
agree & agreeing & agrees & agreed & agreed \\
realize & realizing & realizes & realized & realized \\
value & valuing & values & valued & valued \\
seem & seeming & seems & seemed & seemed \\
hear & hearing & hears & heard & heard \\
doubt & doubting & doubts & doubted & doubted \\
consist & consisting & consists & consisted & consisted \\
smell & smelling & smells & smelled & smelled \\
        \hline
    \end{tabular}}
    \caption{Full list of stative verbs.}
    \label{tab:verb_forms}
\end{table}

\subsection{Threshold Values}
Due to the differences in image and text styles across datasets caused by their respective domains, we determined the threshold values for text similarity and image similarity for each dataset through manual inspection and empirical tuning, as shown in the \autoref{tab:threshold_value}.
For example, in FlintstonesSV, for text similarity, two texts are considered dissimilar if their BERTScore precision and recall are both below 0.98, and their F1 score is below 0.96. Similarly, for image similarity, two images are considered dissimilar if their CLIP similarity score is below 0.94.
\begin{table}[t]
\centering
\resizebox{0.48\textwidth}{!}{
\begin{tabular}{@{}ccccc@{}}
\toprule
              & \multicolumn{3}{c}{BERTScore}                       & CLIP Similarity \\ \midrule
              & precision       & recall          & f1              &                 \\
FlintstonesSV & \textless{}0.98 & \textless{}0.98 & \textless{}0.96 & \textless{}0.94 \\
PororoSV      & \textless{}0.96 & \textless{}0.96 & \textless{}0.95 & \textless{}0.90 \\
VWP           & \textless{}0.98 & \textless{}0.98 & \textless{}0.97 & \textless{}0.95 \\
VIST          & \textless{}0.92 & \textless{}0.92 & \textless{}0.90 & \textless{}0.88 \\ \bottomrule
\end{tabular}}
\caption{The similarity threshold values used in data filtering. }
\label{tab:threshold_value}
\end{table}

\subsection{Prompt used to extract events}\label{append:ev-ext}
The main difference between extracted events and original captions lies in the level of detail: the former contains only the essential event information, such as who did what; the latter, are written by human annotators as complete stories/narratives. Original captions often contain temporal connectives such as "before" "then" or "finally" which may enable models to infer multi-event relations and perform sentence ordering based solely on textual cues.

The prompt we used to extract the events from the original captions by GPT-4 is as follows:

\begin{tcolorbox}[colback=gray!20, colframe=black]
Given a caption: [the original caption here]. Extract a single, concise and clear event sentence from the provided caption. Ensure the returned sentence satisfies the following criteria: (1) The sentence should contain the event itself without phrases such as "the event is" or "event:". (2) If there are multiple events, extract only a single major event. (3) The sentence must contain only one clause with only one verb. (4) The event should be expressed in simple past tense. (5) If no event is detected, return 'NO\_EVENT'.
\end{tcolorbox}

\subsection{Data Statistics after Each Filtering Step}
In \autoref{tab:filtering_stat}, we show the statistics of image sequences left after each data filtering step.
\begin{table}[h]
\centering
\resizebox{0.48\textwidth}{!}{
\begin{tabular}{@{}ccccc@{}}
\toprule
& FlintstonesSV & PororoSV & VWP    & VIST   \\ \midrule
Original image sequences                                                            & 24,433        & 11,444   & 12,627 & 49,700 \\
No stative verbs                                                                    & 10,378        & 3,114    & 1,995  & 20,294 \\
No starting pronoun                                                                 & 10,105        & 2,952    & 914    & 12,216 \\
No similar text                                                                     & 3,092         & 1,952    & 815    & 2,906  \\
No similar image                                                                    & 636           & 644      & 809    & 2,315  \\
No ambiguous image-text                                                             & 633           & 535      & 686    & 2,284  \\
No repetitive image sequences                                                       & 633           & 535      & 498    & 1,880  \\
Without No\_EVENT                                                                   & 612           & 535      & 417    & 1,292  \\
\begin{tabular}[c]{@{}c@{}}Final image sequences \\ after manual check\end{tabular} & 565           & 395      & 316    & 809    \\
Final two-event groups                                                          & 2104          & 864      & 850    & 3742   \\
Final three-event groups                                                        & 916           & 172      & 208    & 830    \\ \bottomrule
\end{tabular}}
\caption{The number of image sequences of each step in data filtering process.}
\label{tab:filtering_stat}
\end{table}

\section{Human Performance Survey}
\label{append:human-survey}
\begin{figure}[ht]
    \centering
    \includegraphics[width=0.48\textwidth]{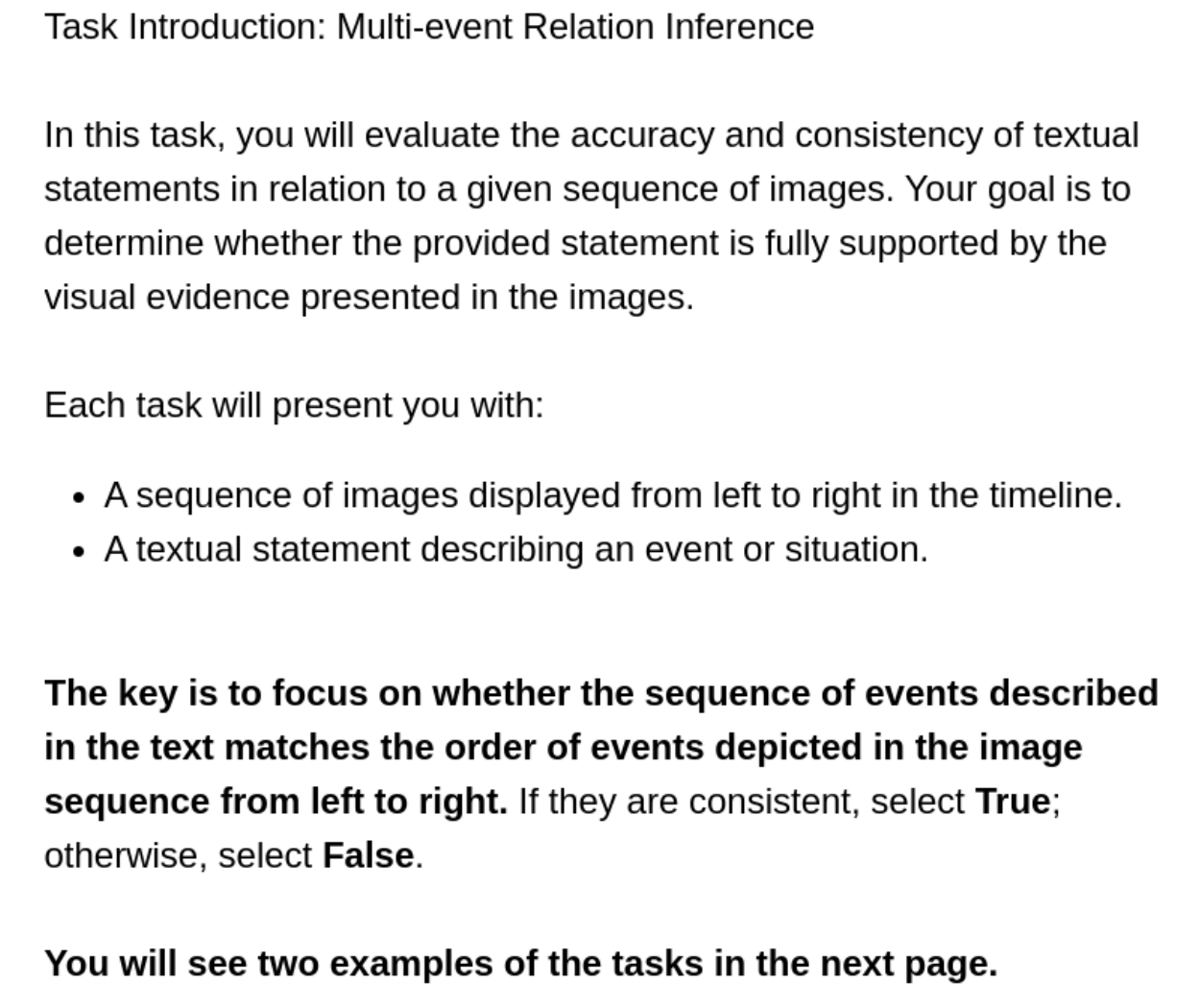}
    \vspace{0.2cm} \includegraphics[width=0.48\textwidth]{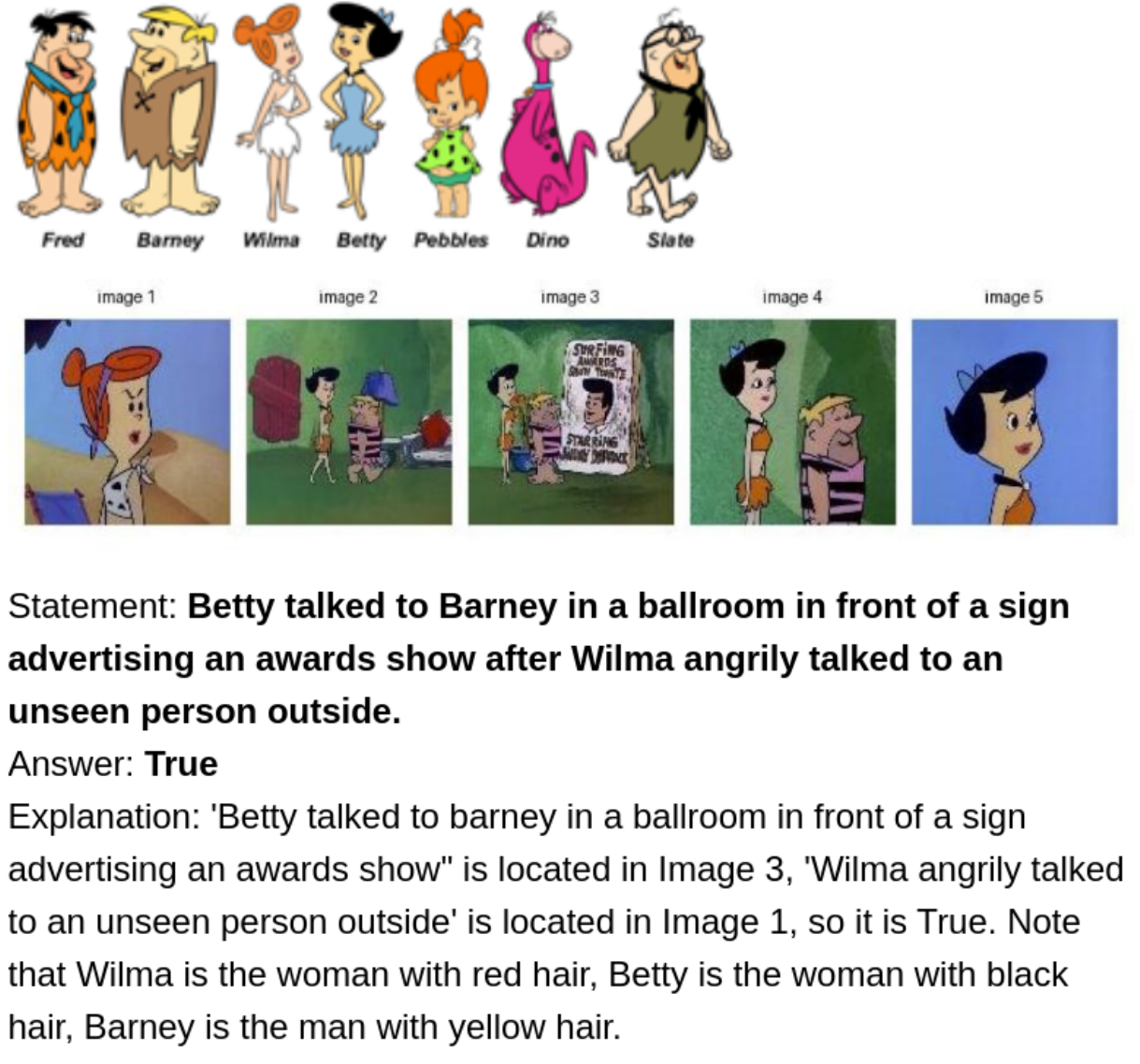}
  
    \caption{An example of task instruction and example question shown in the beginning of human questionnaire}
\label{fig:task_instruct}
\end{figure}

\begin{figure}[ht]
    \centering
    \includegraphics[width=0.5\textwidth]{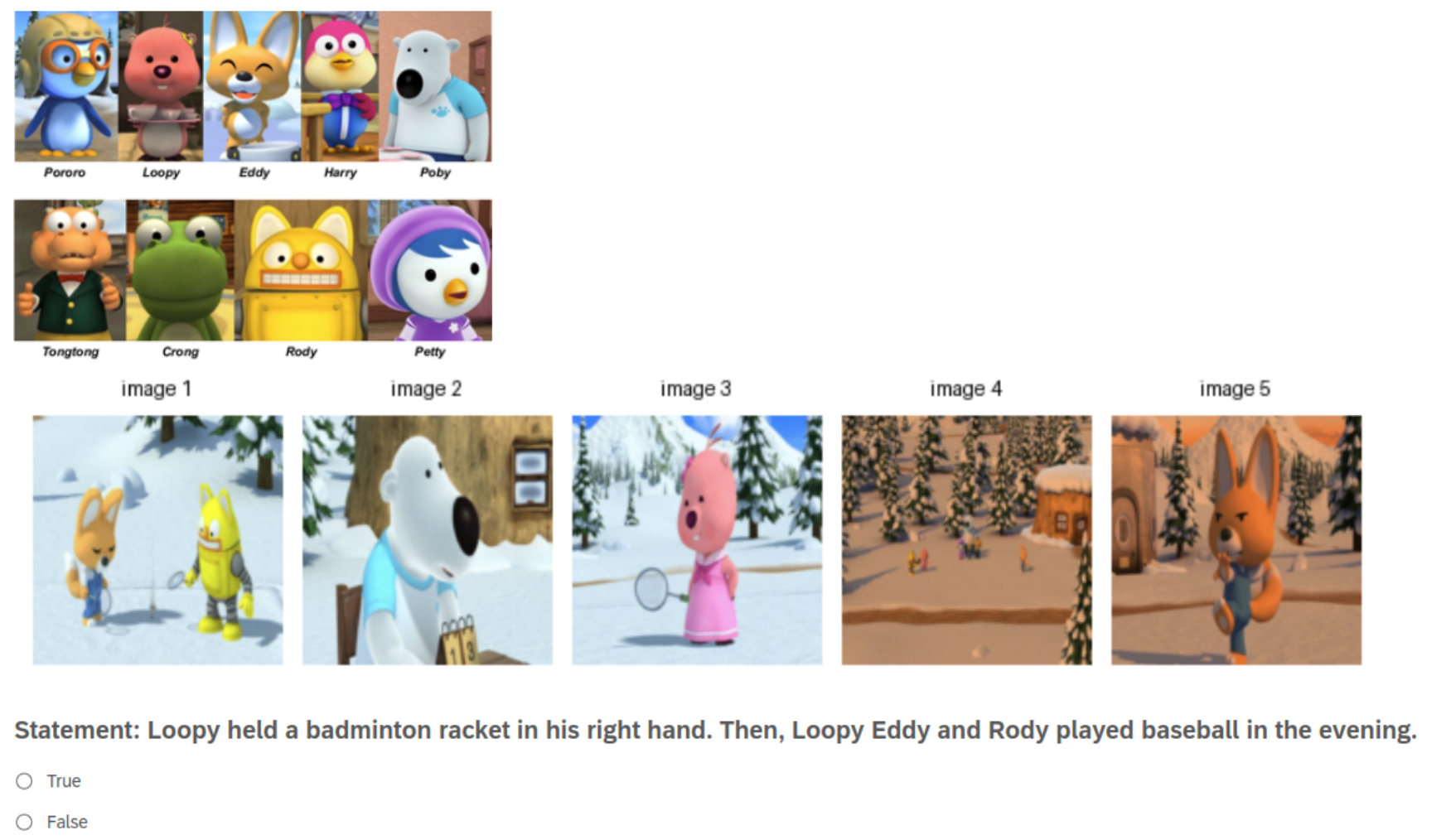}
    \vspace{0.2cm} \includegraphics[width=0.5\textwidth]{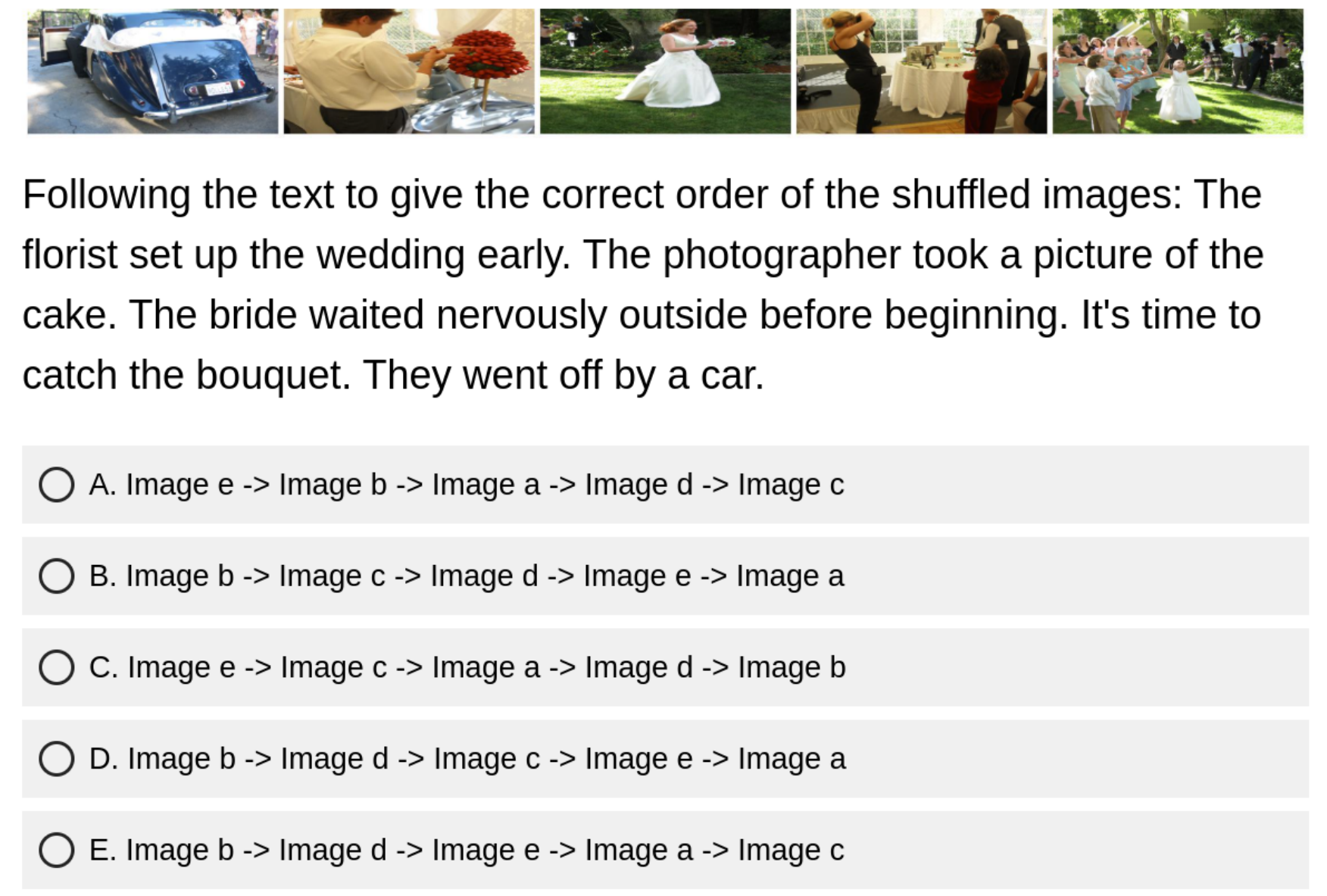}

    \caption{Example question interface used by the human annotators }
\label{fig:example_questions}
\end{figure}

We designed three questionnaires for the human performance survey, corresponding to the three main tasks: event relation inference, sentence ordering, and image ordering.
To ensure that participants fully understand the tasks, we provided task instructions and two sample questions at the beginning of each questionnaire (as shown in \autoref{fig:task_instruct}). Additionally, \autoref{fig:example_questions} presents sample questions that participants were required to answer.
We randomly selected 280 image sequences from TempVS benchmark and collected three responses for each sequence from different annotators.
Participants in MT1 were required to complete 30 questions, while participants in the other two ordering tasks completed 20 questions each. The median completion time was approximately 20 minutes, ensuring that long-time focus did not negatively impact participants' judgment.
We recruited 36 annotators (18 females, 18 males) via Prolific at a hourly rate of £17.1, all of whom were proficient in English and had at least a college-level education. 

\newpage

\section{Prompts}
\label{append:prompts}

\paragraph{Prompt Template} The prompt examples we used for each task are shown below. Our prompt consists of Character Description (only for image sequences from FlintstonesVS and PororoVS), Task Instruction,  Question Text, Response Format and Options (only for main tests), and a prefix indicating the beginning of the answer.

\begin{tcolorbox}[colback=gray!20, colframe=black]
\textbf{GT: Grounding Test }

\textbf{User:} 

\textbf{[Character Description]} Description of character appearance in the images: Fred is chubby, has black hair, a large nose and wears an orange and black spotted short-sleeved loincloth with a blue scarf. 

\textbf{[Task Instruction]} Pick the image from the image sequence that accurately represents the event. When making the choice, focus on the evidence presented in the sequence of images from left to right. 

\textbf{[Question Text]} The event is: Fred was sitting in a room on a stool.

\textbf{[Response Format]} Submit the right number of the image in the sequence as your answer only without additional reasoning or repetition of the instructions.

The answer is:
\end{tcolorbox}

\begin{tcolorbox}[colback=gray!20, colframe=black]
\textbf{MT1: Event Relation Inference }

\textbf{[Character Description]} Description of character appearance in the images: Pororo is a gentoo penguin with an orange-yellow beak wearing a tan aviator's helmet and orange goggles.

\textbf{[Task Instruction]} Is the statement completely accurate and consistent with the content in the sequence of images? When making the choice, focus on the evidence presented in the sequence of images from left to right. 

\textbf{[Question Text]} The statement is: A snowball dropped down a small snow hill after Pororo threw a snowball.

\textbf{[Response Format]} Submit only the right option letter as your answer, e.g., Option [Letter].

\textbf{[Options]} Options are: A. True; B. False.
The answer is:
\end{tcolorbox}

\begin{tcolorbox}[colback=gray!20, colframe=black]

\textbf{MT2: Sentence Ordering}

\textbf{[Task Instruction]}  
You are given an ordered image sequence and several sentences in a random order. Your task is to analyze the content of the sequence of images from left to right and rearrange the sentences into the correct chronological or logical order. Read the image sequence from left to right. Use the images' content to guide your sentence ordering. Avoid assumptions not supported by the image sequence or sentences.

\textbf{[Question Text]} 
The shuffled sentences are:

Sentence a: John prepares to shoot again and fires.

Sentence b: The shadowy figure is hit with the rifle blast.

Sentence c: John takes aim at a shadowy figure up above.

Sentence d: Ken stops the horses and prepares to leave the stagecoach.

Sentence e: Ken is approaching a house out on the prairie on his stagecoach with his two horses.

\textbf{[Response Format]} 
Do not provide explanations or repeat the prompt. Select from the following options and your answer should only be in the format: Option [Letter].

\textbf{[Options]} 

A. Sentence c -> Sentence b -> Sentence a -> Sentence e -> Sentence d;

B. Sentence b -> Sentence e -> Sentence d -> Sentence a -> Sentence c;

C. Sentence e -> Sentence b -> Sentence c -> Sentence d -> Sentence a;

D. Sentence c -> Sentence a -> Sentence e -> Sentence d -> Sentence b;

E. Sentence d -> Sentence c -> Sentence e -> Sentence b -> Sentence a.

The answer is:
\end{tcolorbox}

\begin{tcolorbox}[colback=gray!20, colframe=black]
\textbf{MT3: Image Ordering}

\textbf{[Task Instruction]} The text narrates a story or event sequence. Use your vision-language reasoning to reorder the images to reflect the narrative structure. Carefully read the provided text. Focus on the events, actions, and details described to reorder the images logically.The images are labeled in order as Image a, Image b, Image c, Image d, and Image e, and so on if there are more photos.

\textbf{[Question Text]} The events are: Some people came for the family gathering today. The girls enjoyed some fruit. We played on the swings. The boys lounged in the chair. Grandpa put his grandson on his knee.

\textbf{[Response Format]} Select from the following options and your answer should be in the format: 'Option [Letter]'. Respond with the correct option only, avoiding any explanations or repetition.

\textbf{[Options]} 

A. Image b -> Image c -> Image d -> Image a -> Image e;

B. Image e -> Image d -> Image c -> Image b -> Image a;

C. Image b -> Image e -> Image d -> Image a -> Image c;

D. Image a -> Image c -> Image e -> Image b -> Image d;

E. Image c -> Image b -> Image a -> Image e -> Image d.

The answer is:
\end{tcolorbox}

\paragraph{Variations of Each Prompt Component}
For each component in the prompt, we generate multiple variations (as listed in \autoref{tab:prompt-components}). By combining different components, we ultimately generate 328 distinct prompts.

\onecolumn
\begin{table}[]
\centering
\resizebox{\textwidth}{!}{
\begin{tabular}{@{}
>{\columncolor[HTML]{FFFFFF}}l 
>{\columncolor[HTML]{FFFFFF}}l 
>{\columncolor[HTML]{FFFFFF}}l @{}}
\toprule
\multicolumn{2}{l}{\cellcolor[HTML]{FFFFFF}Task Instruction - definition}                                                                                                                                                                                                                                                                                                                                                                                                                                                                                                                                                                                                                                                                                                                                                                                                                                                                                                                                                                                                                                                                                                                                                                                                                                                                                                                                                                                                                                                                                                                                                                                                         &  \\ \midrule
MT1 & \begin{tabular}[c]{@{}l@{}}- Is the statement completely accurate and consistent with the content in the sequence of images?\\ - Analyze the provided image sequence to determine whether the following statement is True or False.\\ - Review the sequential images and decide whether the provided statement is True or False.\\ - Is the statement entirely consistent and supported by the images in the sequence?\\ - Examine the visual evidence in the provided sequence of images to determine whether the statement is True or False.\\ - Does the statement fully match the information presented in the ordered images? Taking the content in the image seqeunce \\    into account, can you decide whether the statement is True or False?   \\ - Given the multiple images provided in order, can you select the correct answer from True or False considering statement?\end{tabular}                                                                                                                                                                                                                                                                                                                                                                                                                                                                                                                                                                                                                                                                                                                                                                         &  \\ \cmidrule(r){1-2}
MT2 & \begin{tabular}[c]{@{}l@{}}- You are given an ordered image sequence and several sentences in a random order. Your task is to analyze the content of the sequence of \\    images from left to right and rearrange the sentences into the correct chronological or logical order.\\ - Here is an left-to-right image seqeunce and some sentences in random order. Your goal is to determine the correct order of the sentences \\    based on the context, events, or details observable in the images. Consider the visual elements in the sequential images to infer the logical or temporal sequence.\\ - The images are presented in a correct order, but the sentences are not. Your task is to reorder the sentences to match the sequence of events or details \\    in the images from left to right.\\ - You are provided with a sequence of images and several randomly ordered sentences. Your task is to: 1. Understand the context of the image sequence; \\   2. Identify how each sentence relates to the image; 3. Rearrange the sentences to form a coherent order.\\ - Using your multimodal understanding and reasoning of the ordered image sequence and the randomly shuffled sentences , arrange the sentences \\    in the correct sequence to match the flow of events in the image seqeunce.\\ - Presented is an image sequence in the order from left to right along with several unordered sentences. Your task is to determine the correct sequence of the sentences \\   by analyzing the context, events, or details depicted in the images. Use the visual elements in the images to deduce the logical or chronological order.\end{tabular} &  \\ \cmidrule(r){1-2}
MT3 & \begin{tabular}[c]{@{}l@{}}- Rearrange the following images in the correct order based on content in the story.\\ - The provided paragraph describes a sequence of events. Arrange the images in the correct chronological order to match the story.\\ - The images are out of order compared to the text. Identify and reorder them to match the described sequence.\\ - The text narrates a story or event sequence. Use your vision-language reasoning to reorder the images to reflect the narrative structure.\\ - Using your understanding of the text and image content, arrange the images in the correct sequence to match the flow of events in the text.\\ - The paragraph describes events in a specific timeline. Use multimodal reasoning to reorder the images in the correct sequence.\end{tabular}                                                                                                                                                                                                                                                                                                                                                                                                                                                                                                                                                                                                                                                                                                                                                                                                                                                         &  \\ \cmidrule(r){1-2}
GT  & \begin{tabular}[c]{@{}l@{}}- Based on the event described, select the image that best matches it from the following options.\\ - Pick the image from the image sequence that accurately represents the event.\\ - Which is the most fitting image for the described event? Pick from the choices below.\\ - If you were to illustrate the event, which picture would you use?\\ - Find the image from the sequence of images that represents the event most precisely.\\ - Identify the picture that best represents the given event.\end{tabular}                                                                                                                                                                                                                                                                                                                                                                                                                                                                                                                                                                                                                                                                                                                                                                                                                                                                                                                                                                                                                                                                                                                          &  \\ \cmidrule(r){1-2}
\multicolumn{2}{l}{\cellcolor[HTML]{FFFFFF}Task Instruction - requirement}                                                                                                                                                                                                                                                                                                                                                                                                                                                                                                                                                                                                                                                                                                                                                                                                                                                                                                                                                                                                                                                                                                                                                                                                                                                                                                                                                                                                                                                                                                                                                                                                        &  \\ \cmidrule(r){1-2}
MT1 & \begin{tabular}[c]{@{}l@{}}- When making the choice, focus on the evidence presented in the sequence of images from left to right.\\ - Choose the correct option based on the content in the sequential images from left to right.\\ - Only use the left-to-right content of the image sequence to inform the decision.\\ - Evaluate the statement strictly based on the information shown in the sequence of images from left to right.\end{tabular}                                                                                                                                                                                                                                                                                                                                                                                                                                                                                                                                                                                                                                                                                                                                                                                                                                                                                                                                                                                                                                                                                                                                                                                                                       &  \\ \cmidrule(r){1-2}
MT2 & \begin{tabular}[c]{@{}l@{}}- Read the image sequence from left to right. Use the images' content to guide your sentence ordering. Avoid assumptions not supported by the image seqeunce or sentences.\\ - Understand images from left to right in the order. Focus on the visual content of the images to determine the correct order of the sentences. \\   Make sure the reordered sentences form a clear and coherent narrative or description.\\ - Follow the sequence of images from left to right and use their content to determine the correct sentence order. Do not rely on assumptions that are not supported by the images or sentences.\\ - Interpret the images in their left-to-right order, focusing on their visual details to arrange the sentences correctly. \\   Ensure that the reordered sentences create a logical and coherent narrative or description.\end{tabular}                                                                                                                                                                                                                                                                                                                                                                                                                                                                                                                                                                                                                                                                                                                                                                              &  \\ \cmidrule(r){1-2}
MT3 & \begin{tabular}[c]{@{}l@{}}- Carefully read the provided text. Focus on the events, actions, and details described to reorder the images logically.\\ - Focus on matching the actions and events shown in the images with the details described in the text.\\ - Identify key events and details from the text and use them to determine the proper order of the images.\\ - Compare the images with the text description, focusing on the sequence of events to arrange unordered images correctly.\end{tabular}                                                                                                                                                                                                                                                                                                                                                                                                                                                                                                                                                                                                                                                                                                                                                                                                                                                                                                                                                                                                                                                                                                                                                          &  \\ \cmidrule(r){1-2}
GT  & \begin{tabular}[c]{@{}l@{}}- Make your choice by considering the evidence shown in the sequence of images from left to right.\\ - Select the correct option using only the content presented in the sequential images from left to right.\\ - Rely on the left-to-right order of the image sequence to guide your decision.\\ - Assess the statement exclusively based on the information depicted in the sequence of images from left to right.\end{tabular}                                                                                                                                                                                                                                                                                                                                                                                                                                                                                                                                                                                                                                                                                                                                                                                                                                                                                                                                                                                                                                                                                                                                                                                                               &  \\ \cmidrule(r){1-2}
\multicolumn{2}{l}{\cellcolor[HTML]{FFFFFF}Response Format}                                                                                                                                                                                                                                                                                                                                                                                                                                                                                                                                                                                                                                                                                                                                                                                                                                                                                                                                                                                                                                                                                                                                                                                                                                                                                                                                                                                                                                                                                                                                                                                                                       &  \\ \cmidrule(r){1-2}
MT1 & \begin{tabular}[c]{@{}l@{}}- No need to give reasoning process. Submit only the right option letter as your answer, e.g., Option {[}Letter{]}.\\ - Do not tell the reasons of your decision. Provide the most suitable choice letter in the format of 'Option {[}Letter{]}' as your response only.\\ - Do not repeat the prompt or include reasons. Only return the correct option letter in the form of 'Option {[}Letter{]}' as your response.\\ - Please exclude explanations in the response. Offer the most proper choice letter in the format of 'Option {[}Letter{]}' as your answer only.\end{tabular}                                                                                                                                                                                                                                                                                                                                                                                                                                                                                                                                                                                                                                                                                                                                                                                                                                                                                                                                                                                                                                                              &  \\ \cmidrule(r){1-2}
MT2 & \begin{tabular}[c]{@{}l@{}}- Do not provide explanations or repeat the prompt. Select from the following options and your answer should only be in the format: Option {[}Letter{]}.\\ - Provide your answer from the following choices only in the format: 'Option {[}Letter{]}' without explanations or repeating the instructions.\\ - Choose the correct option from the choices provided below and output your answer only as 'Option {[}Letter{]}', avoiding any explanations or repetition.\end{tabular}                                                                                                                                                                                                                                                                                                                                                                                                                                                                                                                                                                                                                                                                                                                                                                                                                                                                                                                                                                                                                                                                                                                                                              &  \\ \cmidrule(r){1-2}
MT3 & \begin{tabular}[c]{@{}l@{}}- Select from the following options and your answer should be in the format: 'Option {[}Letter{]}'. Respond with the correct option only, avoiding any explanations or repetition.\\ - Do not provide explanations or restate the question. Provide your answer from the following choices only in the format: Option {[}Letter{]}.\\ - Choose the correct option from the choices provided below and submit your answer only as 'Option {[}Letter{]}', without any justifications or repetition of the prompt.\end{tabular}                                                                                                                                                                                                                                                                                                                                                                                                                                                                                                                                                                                                                                                                                                                                                                                                                                                                                                                                                                                                                                                                                                                     &  \\ \cmidrule(r){1-2}
GT  & \begin{tabular}[c]{@{}l@{}}- Submit the right number of the image in the sequence as your answer only without additional reasoning or repetition of the instructions.\\ - Provide  only the most suitable image number as your response, avoiding any explanations or repetition.\\ - Only return the correct image number in the provided sequence without additional reasoning details or repetition of the instructions.\end{tabular}                                                                                                                                                                                                                                                                                                                                                                                                                                                                                                                                                                                                                                                                                                                                                                                                                                                                                                                                                                                                                                                                                                                                                                                                                                    &  \\ \bottomrule
\end{tabular}}
\caption{All variations of different components we used to generate the prompts.}
\label{tab:prompt-components}
\end{table}

\paragraph{Chain-of-Thought Prompts}
\label{append:cot-prompt}

\begin{longtable}{|p{2cm}|p{13cm}|}
    \hline
    \textbf{Task} & \textbf{CoT Prompt} \\ 
    \hline
    \endfirsthead
    \multicolumn{2}{c}{{\textit{(Continued from previous page)}}} \\ 
    \hline
    \textbf{List Name} & \textbf{COT Prompt} \\ 
    \hline
    \endhead
    \hline
    \endfoot

    \textbf{MT1} & Analyze the provided image sequence to determine whether the following statement is True or False. When making the choice, carefully examine the evidence presented in the sequence of images from left to right. First, describe the key details and changes observed in each image. Then, explain how these details support or contradict the given statement. Finally, based on your step-by-step reasoning, conclude whether the statement is True or False. Ensure your response follows this format: the reasoning process should be enclosed within \texttt{<think>} and \texttt{</think>} tags, and the final answer should be enclosed within \texttt{<answer>} and \texttt{</answer>} tags. \\ \hline

    \textbf{MT2} & You are presented with an ordered image sequence and several sentences in random order. Your task is to determine the correct order of the sentences based on the context, events, or details observable in the images. Multiple-choice options are provided, with each option representing a possible sequence of the sentences. Think step by step, using the content of the images to guide your reasoning. Avoid assumptions not supported by the image sequence or the sentences. \newline Format your response as follows: Enclose your step-by-step reasoning process within \texttt{<think>} and \texttt{</think>} tags. Enclose the selected option (e.g., Option A, Option B) within \texttt{<answer>} and \texttt{</answer>} tags. \\ \hline

    \textbf{MT3} & Order the following images in the correct sequence based on the content of the story. Compare each image with the text description, carefully analyzing the sequence of events to determine the proper order of the unordered images. \newline Response should be in this format: \newline \texttt{<think>} First, examine the text description to identify key events and their chronological order. Next, analyze each image to match it with the corresponding event described in the text. Consider visual cues, actions, and details in the images that indicate the progression of the story. Arrange the images accordingly to reflect the correct sequence of events. \texttt{</think>} \newline \texttt{<answer>} Option [Your Choice] \texttt{</answer>} \\ \hline

\caption{Chain-of-Thought prompts for different tasks.}
\label{tab:cot_prompts}
\end{longtable}

\section{Experiments}

\subsection{Implementation Details}
\label{append:implem}
We evaluate 38 existing MLLMs for multi-event temporal grounding and reasoning, including both proprietary and open-source models. All open-source models are assessed using their official pre-trained versions available on HuggingFace. 
Detailed configurations of the open-source models evaluated are listed in \autoref{tab:model-details}.
To minimize randomness, we set the temperature to 0.
Experiments on open-source models are conducted using a single NVIDIA H100 GPU. For models exceeding 70B parameters, we adopt the mixed precision (FP16), otherwise the full precision.

Following \citet{Jiang2024MANTISIM}, \citet{DBLP:conf/acl/WangZLLXHYLLBBY24}, \citet{meng2025mmiu} and \citet{wang2025muirbench}, we merge sequential images into a single input, separating them with a thin white band and labeling each with an index (e.g., ``Image 1'' when images are sequentially ordered, or ``Image a'' when images are shuffled). In the preliminary experiments, we tested sequentially inputting images one at a time but observed minimal performance differences compared to merging them as shown in \autoref{tab:sep-results}. The usage of a combined image was alsp motivated by practical constraints in current MLLMs, many of which support only a single image input. Moreover, due to limited context length in some models, inputting multiple images at once can result in token truncation and information loss \citep{dingjie2024milebench}. In the released TempVS benchmark and evaluation toolkit, we provide both the individual image inputs and also the combined multi-image inputs, to allow people to use our benchmark to evaluate the models as they want.

\begin{table}[htbp]
\resizebox{\textwidth}{!}{
\begin{tabular}{@{}ccccccccc@{}}
\toprule
                            & \multicolumn{2}{l}{Two-event Relation Inference} & \multicolumn{2}{l}{Three-event Relation Inference} & \multicolumn{2}{l}{Sentence Ordering- event} & \multicolumn{2}{l}{Image Ordering - event} \\ \midrule
                            & MT                  & MT|GT$_{strict}$              & MT                   & MT|GT$_{strict}$               & MT                & MT|GT$_{strict}$            & MT               & MT|GT$_{strict}$           \\
InternVL2\_5-26B-MPO        & 0.5962              & 0.6053                     & 0.6089               & 0.633                       & 0.6557            & 0.8867                   & 0.3203           & 0.392                   \\
InternVL2\_5-78B-MPO        & 0.5972              & 0.5811                     & 0.5948               & 0.6013                      & 0.7378            & 0.9543                   & 0.3697           & 0.4527                  \\
llava-interleave-qwen-7b & 0.5132              & 0.5325                     & 0.5149               & 0.521                       & 0.2807            & 0.132                    & 0.2083           & 0.1459                  \\
Mantis-8B-Idefics2          & 0.5052              & 0.5154                     & 0.5111               & 0.5377                      & 0.2217            & 0                        & 0.1868           & 0                       \\
LLaVA-NeXT-Video-34B-DPO & 0.5412              & 0.5303                     & 0.5166               & 0.5524                      & 0.2953            & 0                        & 0.1975           & 0.25                    \\
LongVA-7B-DPO               & 0.5479              & 0.5437                     & 0.5257               & 0.5508                      & 0.3371            & 0.75                     & 0.1912           & 0                       \\ \bottomrule
\end{tabular}}
\caption{Results of 6 MLLMs when multiple images are input separately into the models.}
\label{tab:sep-results}
\end{table}

\begin{table}[htbp]
\centering
\resizebox{\linewidth}{!}{
\begin{tabular}{@{}
>{\columncolor[HTML]{FFFFFF}}l 
>{\columncolor[HTML]{FFFFFF}}l 
>{\columncolor[HTML]{FFFFFF}}l 
    >{\columncolor[HTML]{FFFFFF}}l @{}}

\toprule
HuggingFace Model ID             & \# Params & Vision Backbone            & Base LLM                 \\ \midrule
deepseek-ai/deepseek-vl2-tiny                & 3.4B      & SigLIP-SO400M-384          & DeepSeekMoE              \\
deepseek-ai/deepseek-vl2-small               & 16.1B     & SigLIP-SO400M-384          & DeepSeekMoE              \\ \midrule
OpenGVLab/InternVL2\_5-1B                  & 0.9B      & InternViT-300M-448px-V2\_5 & Qwen2.5-0.5B-Instruct    \\
OpenGVLab/InternVL2\_5-1B-MPO              & 0.9B      & InternViT-300M-448px-V2\_5 & Qwen2.5-0.5B-Instruct    \\
OpenGVLab/InternVL2\_5-8B                  & 8.1B      & InternViT-300M-448px-V2\_5 & internlm2\_5-7b-chat     \\
OpenGVLab/InternVL2\_5-8B-MPO              & 8.1B      & InternViT-300M-448px-V2\_5 & internlm2\_5-7b-chat     \\
OpenGVLab/InternVL2\_5-26B                 & 25.5B     & InternViT-6B-448px-V2\_5   & internlm2\_5-20b-chat    \\
OpenGVLab/InternVL2\_5-26B-MPO             & 25.5B     & InternViT-6B-448px-V2\_5   & internlm2\_5-20b-chat    \\
OpenGVLab/InternVL2\_5-78B                 & 78.4B     & InternViT-6B-448px-V2\_5   & Qwen2.5-72B-Instruct     \\
OpenGVLab/InternVL2\_5-78B-MPO             & 78.4B     & InternViT-6B-448px-V2\_5   & Qwen2.5-72B-Instruct     \\ \midrule
deepseek-ai/Janus-Pro-1B                     & 1.0B      & ViT-L-16-SigLIP-384        & DeepSeek-LLM-1.5b-base   \\
deepseek-ai/Janus-Pro-7B                     & 7.0B      & ViT-L-16-SigLIP-384        & DeepSeek-LLM-7b-base     \\ \midrule
llava-hf/llava-interleave-qwen-0.5b-hf    & 0.9B      & SigLIP-400M                & Qwen1.5-0.5B-Chat        \\
llava-hf/llava-interleave-qwen-7b-hf      & 8.1B      & SigLIP-400M                & Qwen1.5-7B-Chat          \\
llava-hf/llava-interleave-qwen-7b-dpo-hf  & 8.1B      & SigLIP-400M                & Qwen1.5-7B-Chat          \\ \midrule
lmms-lab/llava-onevision-qwen2-0.5b-ov    & 0.9B      & siglip-so400m-patch14-384  & Qwen2-0.5B               \\
lmms-lab/llava-onevision-qwen2-0.5b-si    & 0.9B      & siglip-so400m-patch14-384  & Qwen2-0.5B               \\
lmms-lab/llava-onevision-qwen2-7b-ov      & 8.0B      & siglip-so400m-patch14-384  & Qwen2-7B                 \\
lmms-lab/llava-onevision-qwen2-7b-si      & 8.0B      & siglip-so400m-patch14-384  & Qwen2-7B                 \\
lmms-lab/llava-onevision-qwen2-72b-ov-sft & 73.2B     & siglip-so400m-patch14-384  & Qwen2-72B                \\
lmms-lab/llava-onevision-qwen2-72b-si     & 73.2B     & siglip-so400m-patch14-384  & Qwen2-72B                \\ \midrule
llava-hf/LLaVA-NeXT-Video-7B-hf           & 7.1B      & SigLIP-400M                & vicuna-7b-v1.5           \\
llava-hf/LLaVA-NeXT-Video-7B-DPO-hf       & 7.1B      & SigLIP-400M                & vicuna-7b-v1.5           \\
llava-hf/LLaVA-NeXT-Video-34B-hf          & 34.8B     & SigLIP-400M                & vicuna-33b-v1.3          \\
llava-hf/LLaVA-NeXT-Video-34B-DPO-hf      & 34.8B     & SigLIP-400M                & vicuna-33b-v1.3          \\ \midrule
lmms-lab/LongVA-7B                        & 7.9B      & SigLIP-400M                & Qwen2-7B-Instruct        \\
lmms-lab/LongVA-7B-DPO                    & 7.9B      & SigLIP-400M                & Qwen2-7B-Instruct        \\ \midrule
TIGER-Lab/Mantis-8B-Idefics2               & 8.4B      & idefics2-8b                & Mistral-7B-v0.1          \\
TIGER-Lab/Mantis-8B-siglip-llama3          & 8.5B      & SigLIP                     & LLaMA-3-8B               \\ \midrule
microsoft/Phi-3-vision-128k-instruct       & 3.8B      & CLIP ViT-L/14              & Phi-3-mini-128k-instruct \\
microsoft/Phi-3.5-vision-instruct          & 4.2B      & CLIP ViT-L/14              & Phi-3.5-mini-instruct    \\ \midrule
Qwen/Qwen2-VL-2B                      & 2.2B      & CLIP ViT-L/14              & Qwen2-1.5B               \\
Qwen/Qwen2-VL-2B-Instruct             & 2.2B      & CLIP ViT-L/14              & Qwen2-1.5B               \\
Qwen/Qwen2-VL-7B                      & 7.6B      & CLIP ViT-L/14              & Qwen2-7B                 \\
Qwen/Qwen2-VL-7B-Instruct             & 7.6B      & CLIP ViT-L/14              & Qwen2-7B                 \\
Qwen/Qwen2-VL-72B                     & 72.7B     & CLIP ViT-L/14              & Qwen2-72B                \\
Qwen/Qwen2-VL-72B-Instruct            & 72.7B     & CLIP ViT-L/14              & Qwen2-72B                \\ \bottomrule
\end{tabular}}
\caption{Details of all evaluated open-source MLLMs: HuggingFace model id, number of parameters, vision backbone, base LLM.}
\label{tab:model-details}
\end{table}

\clearpage
\subsection{Evaluation Results of 38 MLLMs}
\label{append:full-results}
In this section, we present the complete quantitative results of 38 state-of-the-art multimodal large language models (MLLMs) on the TempVS benchmark. The results for the MT1 event relation reasoning task are shown in \autoref{tab:full-mt1}, while \autoref{tab:full-mt2} presents the results for the MT2 sentence ordering task and \autoref{tab:full-mt3} reports the results for the MT3 image ordering task.
It is important to note that GT represents the overall grounding evaluation, which measures the number of events correctly matched to their corresponding images across the entire benchmark. GT$_{strict}$ denotes the strict grounding evaluation, which calculates the number of image sequences in which every event within a sequence is correctly matched to its corresponding image.

\begin{table}[htbp]
\centering
\resizebox{\textwidth}{!}{
\begin{tabular}{@{}lcccccccc@{}}
\toprule
\multicolumn{1}{l}{}             & \multicolumn{4}{c}{Two-event Relation Inference}    & \multicolumn{4}{c}{Three-event Relation Inference}  \\
\multicolumn{1}{l}{}             & GT     & GT$_{strict}$ & MT     & MT|GT$_{strict}$ & GT     & GT$_{strict}$ & MT     & MT|GT$_{strict}$ \\ \midrule
\rowcolor[HTML]{C0C0C0} 
deepseek-vl2-tiny                & 0.4159 & 0.2082     & 0.4971 & 0.4965        & 0.4292 & 0.1052     & 0.4976 & 0.5064        \\
\rowcolor[HTML]{C0C0C0} 
deepseek-vl2-small               & 0.4116 & 0.1416     & 0.4311 & 0.4223        & 0.4306 & 0.0669     & 0.4436 & 0.4435        \\
InternVL2\_5-1B                  & 0.3636 & 0.1702     & 0.4102 & 0.4073        & 0.3896 & 0.085      & 0.3895 & 0.4270        \\
InternVL2\_5-1B-MPO              & 0.3879 & 0.1881     & 0.3727 & 0.3709        & 0.4002 & 0.0953     & 0.3736 & 0.3569        \\
InternVL2\_5-8B                  & 0.5655 & 0.3976     & 0.5426 & 0.5540        & 0.6246 & 0.3278     & 0.5438 & 0.5556        \\
InternVL2\_5-8B-MPO              & 0.6230 & 0.4697     & 0.5619 & 0.5736        & 0.6871 & 0.4118     & 0.558  & 0.5714        \\
InternVL2\_5-26B                 & 0.6181 & 0.4604     & 0.5705 & 0.5843        & 0.6824 & 0.3958     & 0.5835 & 0.6022        \\
InternVL2\_5-26B-MPO             & 0.6521 & 0.5132     & 0.6032 & 0.6212        & 0.7096 & 0.4595     & 0.6212 & 0.6474        \\
InternVL2\_5-78B                 & 0.6555 & 0.5157     & 0.5423 & 0.5532        & 0.7154 & 0.4733     & 0.5679 & 0.5698        \\
InternVL2\_5-78B-MPO             & 0.7046 & 0.5881     & 0.5847 & 0.5992        & 0.7771 & 0.5653     & 0.6139 & 0.6258        \\
\rowcolor[HTML]{C0C0C0} 
Janus-Pro-1B                     & 0.2359 & 0.0273     & 0.4827 & 0.4814        & 0.2558 & 0.0073     & 0.4645 & 0.4259        \\
\rowcolor[HTML]{C0C0C0} 
Janus-Pro-7B                     & 0.2399 & 0.0429     & 0.3509 & 0.3406        & 0.2555 & 0.0038     & 0.3287 & 0.3929        \\
llava-interleave-qwen-0.5b-hf    & 0.2172 & 0.0245     & 0.4976 & 0.4779        & 0.2182 & 0.0019     & 0.5036 & 0.5012        \\
llava-interleave-qwen-7b-hf      & 0.3521 & 0.1300     & 0.5161 & 0.5239        & 0.4112 & 0.0723     & 0.5007 & 0.4981        \\
llava-interleave-qwen-7b-dpo-hf  & 0.3531 & 0.1358     & 0.5198 & 0.5357        & 0.4102 & 0.0845     & 0.5173 & 0.5319        \\
\rowcolor[HTML]{C0C0C0} 
llava-onevision-qwen2-0.5b-ov    & 0.304  & 0.0859     & 0.4528 & 0.4545        & 0.3343 & 0.0286     & 0.4807 & 0.4764        \\
\rowcolor[HTML]{C0C0C0} 
llava-onevision-qwen2-0.5b-si    & 0.2153 & 0.0235     & 0.4496 & 0.4429        & 0.2156 & 0.0013     & 0.3587 & 0.3731        \\
\rowcolor[HTML]{C0C0C0} 
llava-onevision-qwen2-7b-ov      & 0.5176 & 0.3278     & 0.5602 & 0.5804        & 0.578  & 0.2604     & 0.5753 & 0.5979        \\
\rowcolor[HTML]{C0C0C0} 
llava-onevision-qwen2-7b-si      & 0.4575 & 0.2505     & 0.5292 & 0.5419        & 0.4895 & 0.1403     & 0.5546 & 0.5654        \\
\rowcolor[HTML]{C0C0C0} 
llava-onevision-qwen2-72b-ov-sft & 0.6215 & 0.4641     & 0.5928 & 0.6213        & 0.6844 & 0.405      & 0.6154 & 0.6349        \\
\rowcolor[HTML]{C0C0C0} 
llava-onevision-qwen2-72b-si     & 0.5418 & 0.3593     & 0.5296 & 0.5393        & 0.6015 & 0.2804     & 0.5248 & 0.5298        \\
LLaVA-NeXT-Video-7B-hf           & 0.2731 & 0.0575     & 0.4603 & 0.4603        & 0.2972 & 0.0135     & 0.4486 & 0.4723        \\
LLaVA-NeXT-Video-7B-DPO-hf       & 0.2718 & 0.0596     & 0.4672 & 0.466         & 0.2981 & 0.014      & 0.4520 & 0.4615        \\
LLaVA-NeXT-Video-34B-hf          & 0.2742 & 0.0652     & 0.5847 & 0.5839        & 0.3053 & 0.0159     & 0.5947 & 0.5763        \\
LLaVA-NeXT-Video-34B-DPO-hf      & 0.3028 & 0.0825     & 0.5330  & 0.5386        & 0.3289 & 0.0205     & 0.5248 & 0.5263        \\
\rowcolor[HTML]{C0C0C0} 
LongVA-7B                        & 0.3015 & 0.0874     & 0.5468 & 0.5613        & 0.3138 & 0.0208     & 0.5596 & 0.6169        \\
\rowcolor[HTML]{C0C0C0} 
LongVA-7B-DPO                    & 0.3241 & 0.1132     & 0.5319 & 0.5582        & 0.3449 & 0.0386     & 0.5233 & 0.5350         \\
Mantis-8B-Idefics2               & 0.3452 & 0.1218     & 0.5193 & 0.533         & 0.3586 & 0.041      & 0.5197 & 0.5164        \\
Mantis-8B-siglip-llama3          & 0.2444 & 0.0443     & 0.5238 & 0.5368        & 0.2392 & 0.0065     & 0.5251 & 0.5833        \\
\rowcolor[HTML]{C0C0C0} 
Phi-3-vision-128k-instruct       & 0.2226 & 0.0379     & 0.5196 & 0.5219        & 0.2332 & 0.0065     & 0.5132 & 0.4583        \\
\rowcolor[HTML]{C0C0C0} 
Phi-3.5-vision-instruct          & 0.2235 & 0.0400     & 0.4904 & 0.4774        & 0.2316 & 0.0078     & 0.4877 & 0.4828        \\
Qwen2-VL-2B                      & 0.3053 & 0.088      & 0.4935 & 0.4842        & 0.3188 & 0.0313     & 0.4717 & 0.4923        \\
Qwen2-VL-2B-Instruct             & 0.3815 & 0.1738     & 0.5271 & 0.5314        & 0.4003 & 0.0777     & 0.5051 & 0.5052        \\
Qwen2-VL-7B                      & 0.4032 & 0.1664     & 0.5122 & 0.4982        & 0.4233 & 0.0899     & 0.4991 & 0.4603        \\
Qwen2-VL-7B-Instruct             & 0.5144 & 0.3239     & 0.5397 & 0.5542        & 0.5415 & 0.2124     & 0.5358 & 0.5534        \\
Qwen2-VL-72B                     & 0.4906 & 0.2997     & 0.5461 & 0.5738        & 0.5167 & 0.1935     & 0.5429 & 0.5900        \\
Qwen2-VL-72B-Instruct            & 0.5078 & 0.3167     & 0.5395 & 0.5643        & 0.5427 & 0.2059     & 0.5563 & 0.6062        \\
\rowcolor[HTML]{C0C0C0} 
GPT-4o                            & 0.7043 & 0.6034     & 0.5827 & 0.6005        & 0.7807 & 0.5704     & 0.6452 & 0.6644        \\ \bottomrule
\end{tabular}}
\caption{Full results for MT1: event relation inference}
\label{tab:full-mt1}
\end{table}

\begin{table}[]
\centering
\resizebox{\textwidth}{!}{
\begin{tabular}{@{}ccccccccc@{}}
\toprule
                                 & \multicolumn{4}{c}{Ordering Sentences (events)} & \multicolumn{4}{c}{Ordering Sentences (captions)} \\
                                 & GT      & GT$_{strict}$  & MT      & MT| GT$_{strict}$ & GT       & GT$_{strict}$  & MT      & MT| GT$_{strict}$  \\ \midrule
\rowcolor[HTML]{C0C0C0} 
deepseek-vl2-tiny                & 0.3701  & 0.0084      & 0.1953  & 0.25          & 0.4415   & 0.0154      & 0.1823  & 0.1786         \\
\rowcolor[HTML]{C0C0C0} 
deepseek-vl2-small               & 0.3021  & 0.0037      & 0.1574  & 0.1429        & 0.3735   & 0.0061      & 0.1707  & 0.1818         \\
InternVL2\_5-1B                  & 0.3462  & 0.0089      & 0.2253  & 0.1176        & 0.3589   & 0.0105      & 0.2037  & 0.2105         \\
InternVL2\_5-1B-MPO              & 0.3636  & 0.0084      & 0.2168  & 0.4375        & 0.3833   & 0.0099      & 0.2032  & 0.3889         \\
InternVL2\_5-8B                  & 0.5294  & 0.0711      & 0.4589  & 0.5778        & 0.5574   & 0.0765      & 0.5424  & 0.6259         \\
InternVL2\_5-8B-MPO              & 0.5747  & 0.0963      & 0.5663  & 0.7377        & 0.6183   & 0.1415      & 0.6289  & 0.7704         \\
InternVL2\_5-26B                 & 0.5765  & 0.1016      & 0.5674  & 0.7358        & 0.6137   & 0.1311      & 0.63    & 0.7689         \\
InternVL2\_5-26B-MPO             & 0.6089  & 0.1258      & 0.6989  & 0.9079        & 0.6491   & 0.1696      & 0.7693  & 0.8734         \\
InternVL2\_5-78B                 & 0.6121  & 0.1363      & 0.6695  & 0.8494        & 0.6536   & 0.1845      & 0.7109  & 0.8388         \\
InternVL2\_5-78B-MPO             & 0.6604  & 0.1842      & 0.7984  & 0.9657        & 0.7064   & 0.2594      & 0.8634  & 0.9639         \\
\rowcolor[HTML]{C0C0C0} 
Janus-Pro-1B                     & 0.2344  & 0           & 0.1858  & -           & 0.2497   & 0.0006      & 0.1982  & 0              \\
\rowcolor[HTML]{C0C0C0} 
Janus-Pro-7B                     & 0.2409  & 0           & 0.1705  & -          & 0.2623   & 0           & 0.1525  & -           \\
llava-interleave-qwen-0.5b-hf    & 0.2161  & 0           & 0.2068  & -          & 0.2269   & 0           & 0.2065  & -           \\
llava-interleave-qwen-7b-hf      & 0.3363  & 0.0032      & 0.2505  & 0.1667        & 0.3477   & 0.0039      & 0.2704  & 0              \\
llava-interleave-qwen-7b-dpo-hf  & 0.3332  & 0.0047      & 0.2679  & 0.3333        & 0.3409   & 0.0099      & 0.2996  & 0.3333         \\
\rowcolor[HTML]{C0C0C0} 
llava-onevision-qwen2-0.5b-ov    & 0.2931  & 0.0005      & 0.1884  & 0             & 0.3118   & 0.0006      & 0.1839  & 0              \\
\rowcolor[HTML]{C0C0C0} 
llava-onevision-qwen2-0.5b-si    & 0.2106  & 0           & 0.1895  & -          & 0.2199   & 0           & 0.1944  & -           \\
\rowcolor[HTML]{C0C0C0} 
llava-onevision-qwen2-7b-ov      & 0.4839  & 0.0447      & 0.4421  & 0.4118        & 0.5118   & 0.0683      & 0.4692  & 0.4758         \\
\rowcolor[HTML]{C0C0C0} 
llava-onevision-qwen2-7b-si      & 0.426   & 0.0168      & 0.4284  & 0.6875        & 0.4119   & 0.0116      & 0.4537  & 0.7619         \\
\rowcolor[HTML]{C0C0C0} 
llava-onevision-qwen2-72b-ov-sft & 0.5838  & 0.0984      & 0.6516  & 0.8182        & 0.6179   & 0.1399      & 0.7506  & 0.8661         \\
\rowcolor[HTML]{C0C0C0} 
llava-onevision-qwen2-72b-si     & 0.5115  & 0.0411      & 0.61    & 0.8077        & 0.5464   & 0.0743      & 0.6691  & 0.7852         \\
LLaVA-NeXT-Video-7B-hf           & 0.2645  & 0           & 0.1895  & -          & 0.2742   & 0           & 0.1823  & -           \\
LLaVA-NeXT-Video-7B-DPO-hf       & 0.2666  & 0           & 0.1963  & -          & 0.2736   & 0           & 0.1938  & -           \\
LLaVA-NeXT-Video-34B-hf          & 0.2646  & 0.0005      & 0.3184  & 1             & 0.2831   & 0           & 0.3343  & -           \\
LLaVA-NeXT-Video-34B-DPO-hf      & 0.2882   & 0.0016      & 0.3095  & 0             & 0.2984   & 0.0006      & 0.3133  & 0              \\
\rowcolor[HTML]{C0C0C0} 
LongVA-7B                        & 0.286   & 0.0016      & 0.3421  & 0.6667        & 0.2909   & 0.0022      & 0.353   & 0.5            \\
\rowcolor[HTML]{C0C0C0} 
LongVA-7B-DPO                    & 0.3089  & 0.0026      & 0.3547  & 0.8           & 0.2997   & 0.0028      & 0.3618  & 0.6            \\
Mantis-8B-Idefics2               & 0.3326  & 0.0011      & 0.2216  & 0             & 0.2868   & 0.0022      & 0.2081  & 0              \\
Mantis-8B-siglip-llama3          & 0.2406  & 0.0005      & 0.2142  & 0             & 0.2461   & 0           & 0.2153  & -           \\
\rowcolor[HTML]{C0C0C0} 
Phi-3-vision-128k-instruct       & 0.2243  & 0           & 0.2316  & -          & 0.2214   & 0           & 0.2329  & -           \\
\rowcolor[HTML]{C0C0C0} 
Phi-3.5-vision-instruct          & 0.2262  & 0           & 0.2311  & -          & 0.2331   & 0           & 0.2539  & -           \\
Qwen2-VL-2B                      & 0.2873  & 0.0004      & 0.1957  & 0             & 0.3274   & 0.0011      & 0.1995  & 0              \\
Qwen2-VL-2B-Instruct             & 0.3638  & 0.0058      & 0.1658  & 0.0909        & 0.333    & 0.0015      & 0.2014  & 0.5            \\
Qwen2-VL-7B                      & 0.4592  & 0.0028      & 0.3097  & 0.25          & 0.4879   & 0.0076      & 0.3276  & 0.5455         \\
Qwen2-VL-7B-Instruct             & 0.4813  & 0.0342      & 0.4253  & 0.6462        & 0.5098   & 0.0441      & 0.446   & 0.6125         \\
Qwen2-VL-72B                     & 0.4539  & 0.0258      & 0.4363  & 0.7143        & 0.4914   & 0.0435      & 0.5319  & 0.7215         \\
Qwen2-VL-72B-Instruct            & 0.4756  & 0.0368      & 0.4632  & 0.6429        & 0.5085   & 0.0496      & 0.5507  & 0.7            \\
\rowcolor[HTML]{C0C0C0} 
GPT-4o                            & 0.6708  & 0.1863      & 0.534   & 0.5389        & 0.7231   & 0.2357      & 0.615   & 0.5532         \\ \bottomrule
\end{tabular}
}
\caption{Full results for MT2: sentence ordering with events/captions.}
\label{tab:full-mt2}
\end{table}

\begin{table}[]
\centering
\begin{tabular}{@{}ccccc@{}}
\toprule
\multicolumn{1}{l}{}             & \multicolumn{2}{l}{Ordering Images (events)}               & \multicolumn{2}{l}{Ordering Images (captions)}             \\
\multicolumn{1}{l}{}             & \multicolumn{1}{l}{MT} & \multicolumn{1}{c}{MT| GT$_{strict}$} & \multicolumn{1}{l}{MT} & \multicolumn{1}{c}{MT| GT$_{strict}$} \\ \midrule
\rowcolor[HTML]{C0C0C0} 
deepseek-vl2-tiny                & 0.2042                 & 0.4286                            & 0.2072                 & 0.125                             \\
\rowcolor[HTML]{C0C0C0} 
deepseek-vl2-small               & 0.1663                 & 0                                 & 0.1553                 & 0                                 \\
InternVL2\_5-1B                  & 0.2107                 & 0.1429                            & 0.1997                 & 0.0625                            \\
InternVL2\_5-1B-MPO              & 0.2052                 & 0.1538                            & 0.2002                 & 0.2667                            \\
InternVL2\_5-8B                  & 0.2766                 & 0.2783                            & 0.2651                 & 0.4113                            \\
InternVL2\_5-8B-MPO              & 0.2986                 & 0.3841                            & 0.318                  & 0.3816                            \\
InternVL2\_5-26B                 & 0.2666                 & 0.2727                            & 0.3165                 & 0.3589                            \\
InternVL2\_5-26B-MPO             & 0.344                  & 0.3971                            & 0.3949                 & 0.4369                            \\
InternVL2\_5-78B                 & 0.311                  & 0.4                               & 0.3849                 & 0.4709                            \\
InternVL2\_5-78B-MPO             & 0.4099                 & 0.4883                            & 0.5382                 & 0.6966                            \\
\rowcolor[HTML]{C0C0C0} 
Janus-Pro-1B                     & 0.2247                 & -                               & 0.2227                 & 0                                 \\
\rowcolor[HTML]{C0C0C0} 
Janus-Pro-7B                     & 0.2092                 & -                               & 0.2097                 & -                               \\
llava-interleave-qwen-0.5b-hf    & 0.2017                 & -                               & 0.2077                 & -                               \\
llava-interleave-qwen-7b-hf      & 0.2087                 & 0.1667                            & 0.1997                 & 0.2222                            \\
llava-interleave-qwen-7b-dpo-hf  & 0.2167                 & 0.3846                            & 0.2127                 & 0.1905                            \\
\rowcolor[HTML]{C0C0C0} 
llava-onevision-qwen2-0.5b-ov    & 0.1942                 & 1                                 & 0.2082                 & 0                                 \\
\rowcolor[HTML]{C0C0C0} 
llava-onevision-qwen2-0.5b-si    & 0.2032                 & -                               & 0.1902                 & -                               \\
\rowcolor[HTML]{C0C0C0} 
llava-onevision-qwen2-7b-ov      & 0.2132                 & 0.1429                            & 0.2157                 & 0.1983                            \\
\rowcolor[HTML]{C0C0C0} 
llava-onevision-qwen2-7b-si      & 0.2077                 & 0.35                              & 0.2047                 & 0.4375                            \\
\rowcolor[HTML]{C0C0C0} 
llava-onevision-qwen2-72b-ov-sft & 0.2756                 & 0.3182                            & 0.2906                 & 0.3647                            \\
\rowcolor[HTML]{C0C0C0} 
llava-onevision-qwen2-72b-si     & 0.2546                 & 0.3553                            & 0.2461                 & 0.2677                            \\
LLaVA-NeXT-Video-7B-hf           & 0.2102                 & -                               & 0.2132                 & -                               \\
LLaVA-NeXT-Video-7B-DPO-hf       & 0.2067                 & -                               & 0.2082                 & -                               \\
LLaVA-NeXT-Video-34B-hf          & 0.1977                 & 0                                 & 0.1997                 & -                               \\
LLaVA-NeXT-Video-34B-DPO-hf      & 0.1887                 & 0.3333                            & 0.1932                 & 0                                 \\
\rowcolor[HTML]{C0C0C0} 
LongVA-7B                        & 0.1952                 & -                               & 0.1902                 & -                               \\
\rowcolor[HTML]{C0C0C0} 
LongVA-7B-DPO                    & 0.1962                 & 1                                 & 0.1937                 & 0                                 \\
Mantis-8B-Idefics2               & 0.1862                 & 0                                 & 0.1922                 & 0.5                               \\
Mantis-8B-siglip-llama3          & 0.2002                 & 0                                 & 0.1987                 & -                               \\
\rowcolor[HTML]{C0C0C0} 
Phi-3-vision-128k-instruct       & 0.1857                 & -                               & 0.1897                 & -                               \\
\rowcolor[HTML]{C0C0C0} 
Phi-3.5-vision-instruct          & 0.1922                 & -                               & 0.1832                 & -                               \\
Qwen2-VL-2B                      & 0.1917                 & -                               & 0.1859                 & 0                                 \\
Qwen2-VL-2B-Instruct             & 0.1438                 & 0.125                             & 0.1483                 & 1                                 \\
Qwen2-VL-7B                      & 0.1917                 & 0                                 & 0.2067                 & 0.1818                            \\
Qwen2-VL-7B-Instruct             & 0.2312                 & 0.2041                            & 0.2456                 & 0.3585                            \\
Qwen2-VL-72B                     & 0.2416                 & 0.3548                            & 0.2406                 & 0.4561                            \\
Qwen2-VL-72B-Instruct            & 0.2651                 & 0.4727                            & 0.2811                 & 0.4737                            \\
\rowcolor[HTML]{C0C0C0} 
GPT-4o                            & 0.2257                 & 0.2353                            & 0.2297                 & 0.2349                            \\ \bottomrule
\end{tabular}
\caption{Full results for MT3: image ordering with events/captions.}
\label{tab:full-mt3}
\end{table}

\newpage
\subsection{Supplementary Results}
\paragraph{Performance of text-only LLMs} The results of the language-only LLMs on our benchmark before and after this filtering step are shown  in \autoref{tab:textonly-remove}. For each LLM, the first row shows results before questions removal, and the second row shows results after. We observe that, after filtering out questions easily answered by language-only LLMs, their performance drops sharply, approaching or falling below random guessing level.

\begin{table}[htbp]
\resizebox{\textwidth}{!}{
\begin{tabular}{@{}ccccc@{}}
\toprule
                                       & Two-event Relation & Three-event Relation & Sentence Ordering - Event & Sentence Ordering - Caption \\ \midrule
\multirow{2}{*}{Phi-3.5-mini-instruct} & 0.5031             & 0.4999               & 0.3257                    & 0.3368                      \\
                                       & 0.1798             & 0.2352               & 0.1048                    & 0.1242                      \\
\multirow{2}{*}{Llama-3.1-8B}          & 0.5023             & 0.5059               & 0.2091                    & 0.1995                      \\
                                       & 0.2767             & 0.2094               & 0.0952                    & 0.1095                      \\
\multirow{2}{*}{Qwen2.5-72B-Instruct}  & 0.5361             & 0.5349               & 0.5260                    & 0.6108                      \\
                                       & 0.2952             & 0.3010               & 0.2722                    & 0.3203                      \\ \bottomrule
\end{tabular}}
\caption{Performance of language-only LLMs before and after filtering out text-answerable questions.  For each LLM, the first row contains the results before removing the questions that can be correctly answered by at least two language-only LLMs, and the second row contains the results after removing these questions.}
\label{tab:textonly-remove}
\end{table}

\paragraph{Results comparison before and after filtering  text-answerable questions} We provide the results of 5 representative MLLMs’ performance before and after this filtering step in \autoref{tab:mllm-remove}. We observe a performance drop across all MLLMs, indicating that our benchmark becomes more challenging after removing questions that can be answered using text alone. This further reinforces the requirement for MLLMs to truly understand the visual content. 

\begin{table}[htbp]
\resizebox{\textwidth}{!}{
\begin{tabular}{@{}ccccc@{}}
\toprule
\multicolumn{1}{l}{}                              & Two-event Relation & Three-event Relation & Sentence Ordering - Event & Sentence Ordering - Caption \\ \midrule
\multirow{2}{*}{InternVL2\_5-26B-MPO}             & 0.6643             & 0.6743               & 0.7410                    & 0.7940                      \\
                                                  & 0.6032             & 0.6212               & 0.6989                    & 0.7693                      \\
\multirow{2}{*}{InternVL2\_5-78B-MPO}             & 0.6755             & 0.6948               & 0.8334                    & 0.9037                      \\
                                                  & 0.5847             & 0.6139               & 0.7984                    & 0.8634                      \\
\multirow{2}{*}{llava-onevision-qwen2-72b-ov-sft} & 0.6712             & 0.6841               & 0.7211                    & 0.8093                      \\
                                                  & 0.5928             & 0.6154               & 0.6516                    & 0.7506                      \\
\multirow{2}{*}{LLaVA-NeXT-Video-34B-hf}          & 0.7158             & 0.6828               & 0.4014                    & 0.3993                      \\
                                                  & 0.5847             & 0.5947               & 0.3184                    & 0.3343                      \\
\multirow{2}{*}{Qwen2-VL-72B-Instruct}            & 0.6505             & 0.6533               & 0.5698                    & 0.6606                      \\
                                                  & 0.5395             & 0.5563               & 0.4632                    & 0.5507                      \\ \bottomrule
\end{tabular}}
\caption{Performance of five MLLMs before and after filtering out text-answerable questions.  For each MLLM, the first row contains the results before removing the questions that can be correctly answered by at least two language-only LLMs, and the second row contains the results after removing these questions.}
\label{tab:mllm-remove}
\end{table}

\paragraph{Performance of image ordering task with images only} We perform image ordering task without text descriptions. As shown in the \autoref{tab:image-only}, even for the best-performing MLLMs, ordering based on images alone remains highly challenging as their performance is near random guessing. This further confirms that our benchmark requires models to jointly understand temporal information from both text and images. Models cannot rely solely on prior knowledge to infer the correct order of images.  This is precisely why we chose visual stories as the data source of our benchmark: accurate temporal reasoning cannot be achieved by relying on a single modality alone.

\renewcommand{\arraystretch}{0.9}
\begin{table}[ht]
\centering
\begin{tabular}{@{}cccc@{}}
\toprule
           & with event & with caption & image only \\ \midrule
InternVL2\_5-26B-MPO      & 0.3440     & 0.3949       & 0.2430     \\
InternVL2\_5-78B-MPO      & 0.4099     & 0.5382       & 0.2709     \\
llava-onevision-qwen2-72b & 0.2756     & 0.2906       & 0.2023     \\ \bottomrule
\end{tabular}
\caption{Performance of three MLLMs in image ordering task with event text, original captions and only images.}
\label{tab:image-only}
\end{table}


\newpage
\section{Additional Examples in TempVS}
\label{append:more-examples}
We provide more examples in TempVS benchmark.

\begin{figure}[htbp]
    \centering
     \includegraphics[width=\linewidth, trim=0 50 0 0, clip]{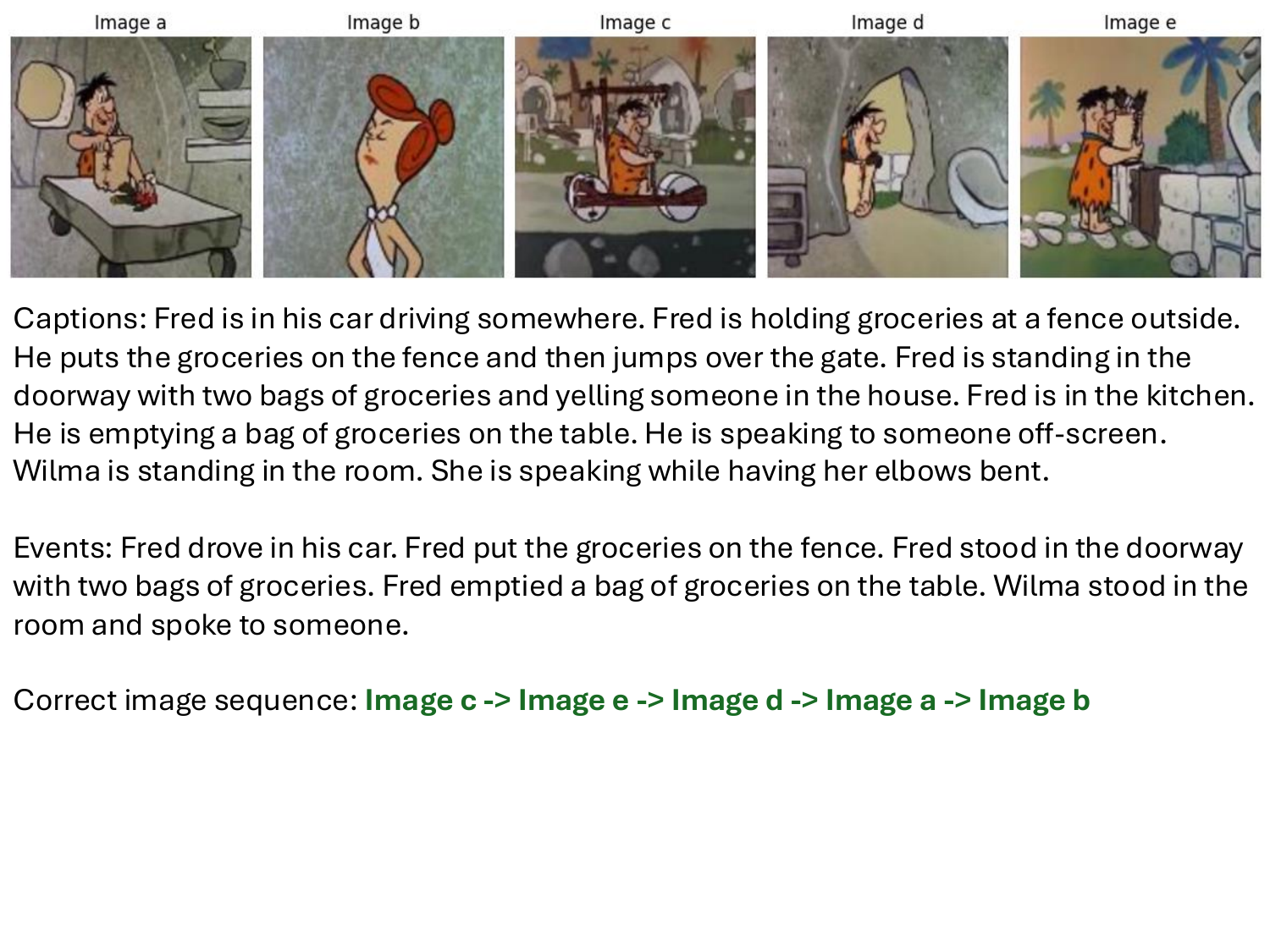}\\
    \includegraphics[width=\linewidth, trim=0 190 0 0, clip]{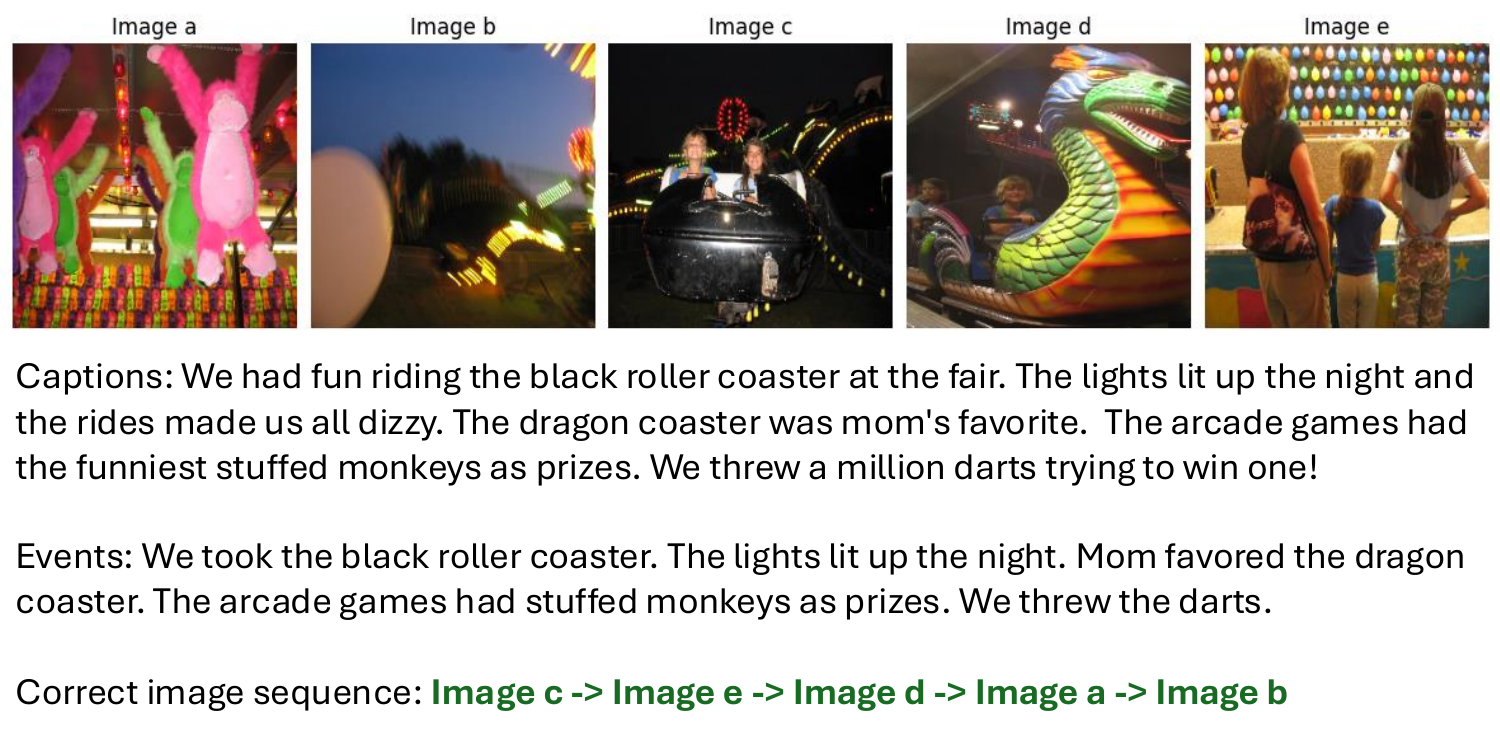}
\end{figure}

\begin{figure}[htbp]
    \centering
    \includegraphics[width=\linewidth]{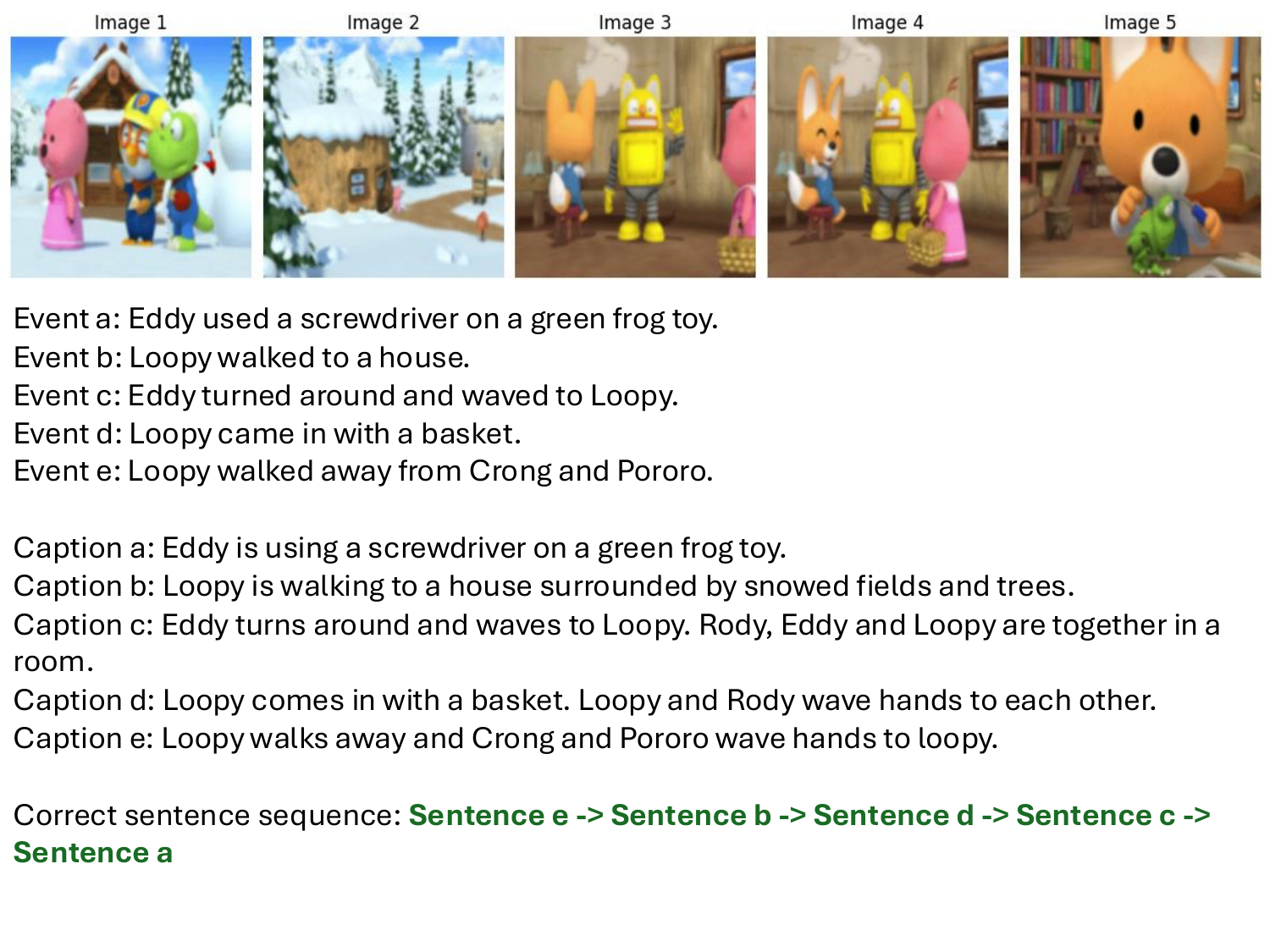} \\
    \includegraphics[width=\linewidth]{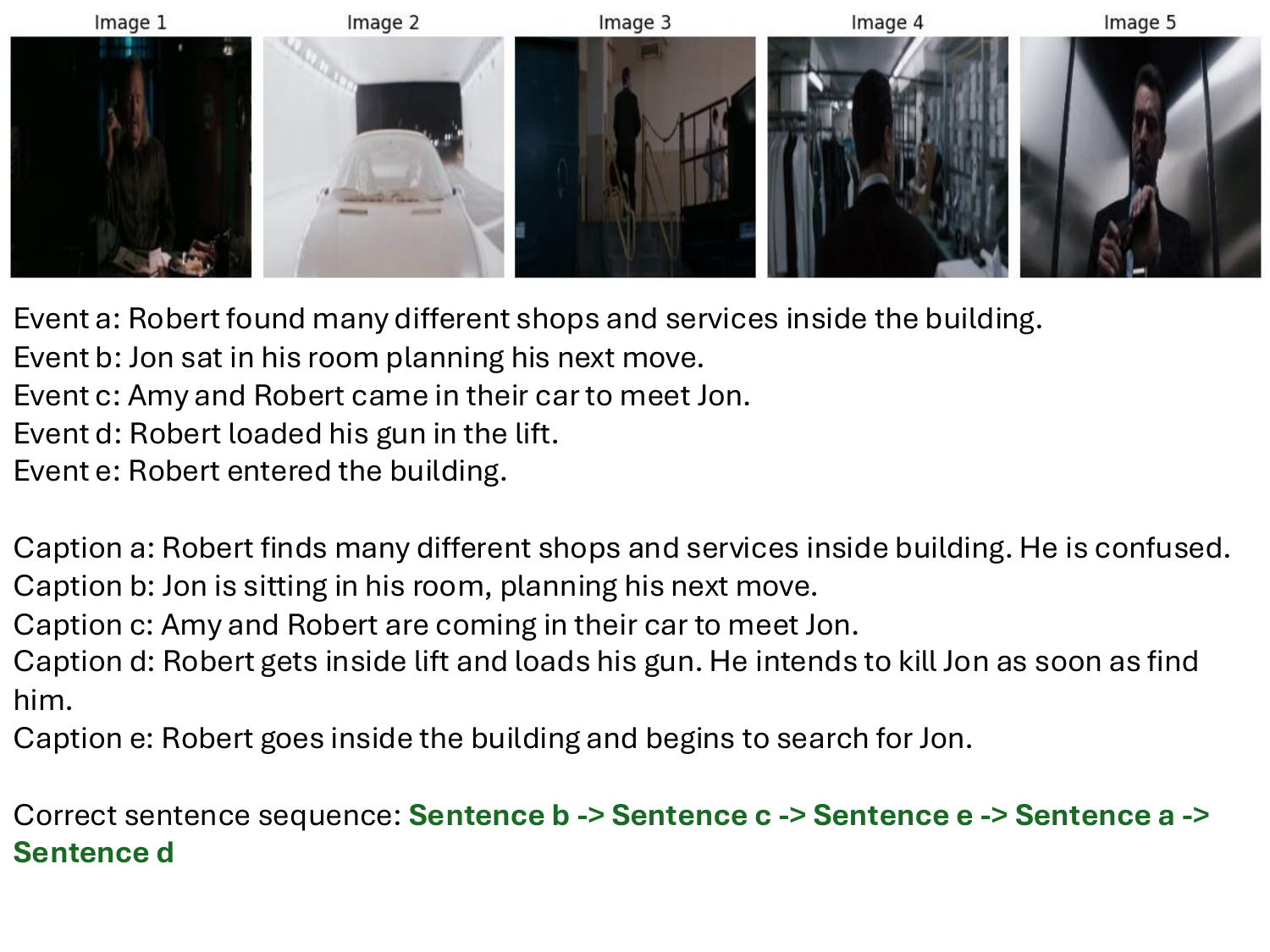} \\
\end{figure}

\begin{figure}[htbp]
    \centering
    \includegraphics[width=\linewidth]{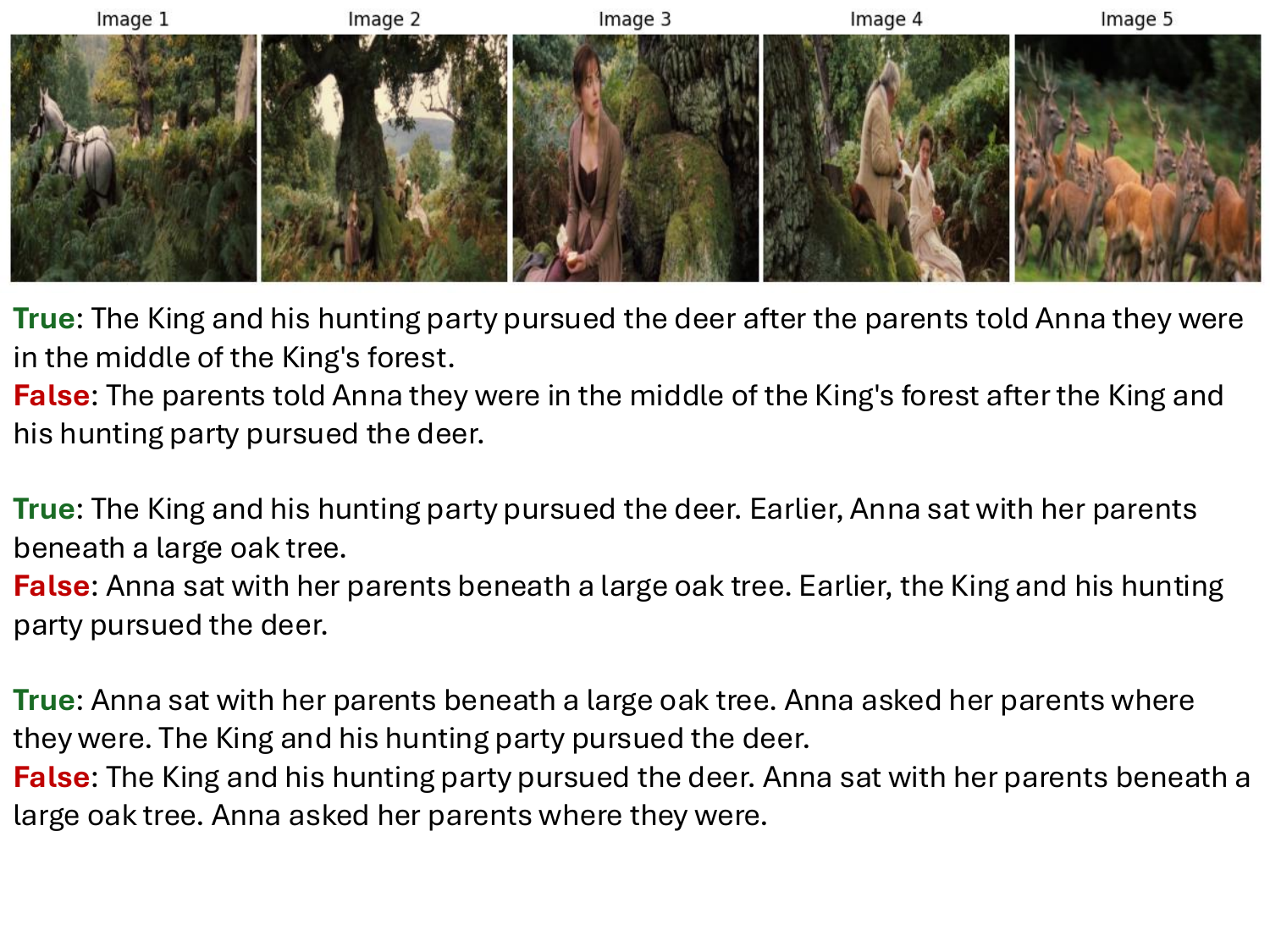} \\
    \includegraphics[width=\linewidth]{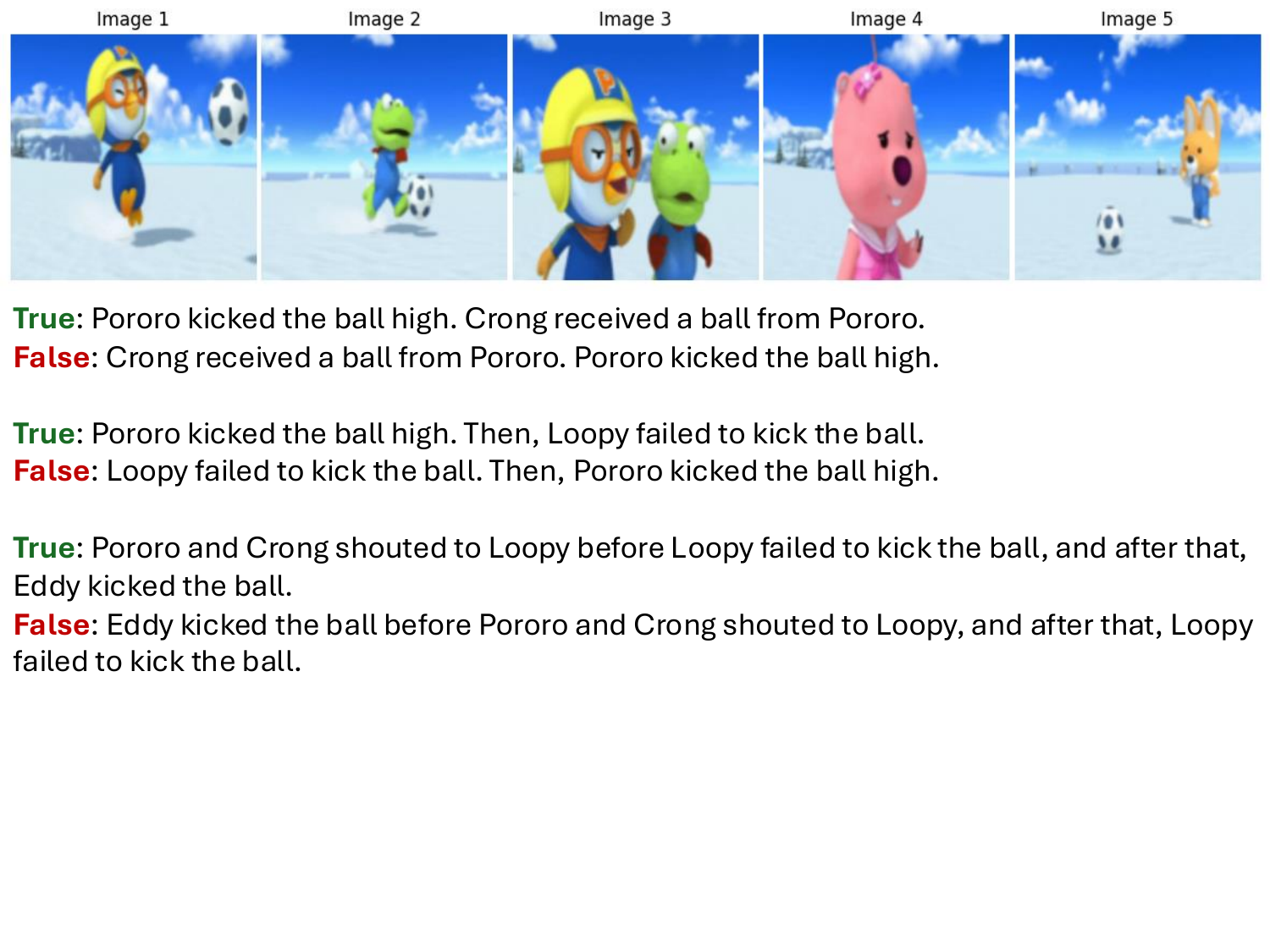} \\
\end{figure}

\begin{figure}[htbp]
    \centering
    \includegraphics[width=\linewidth]{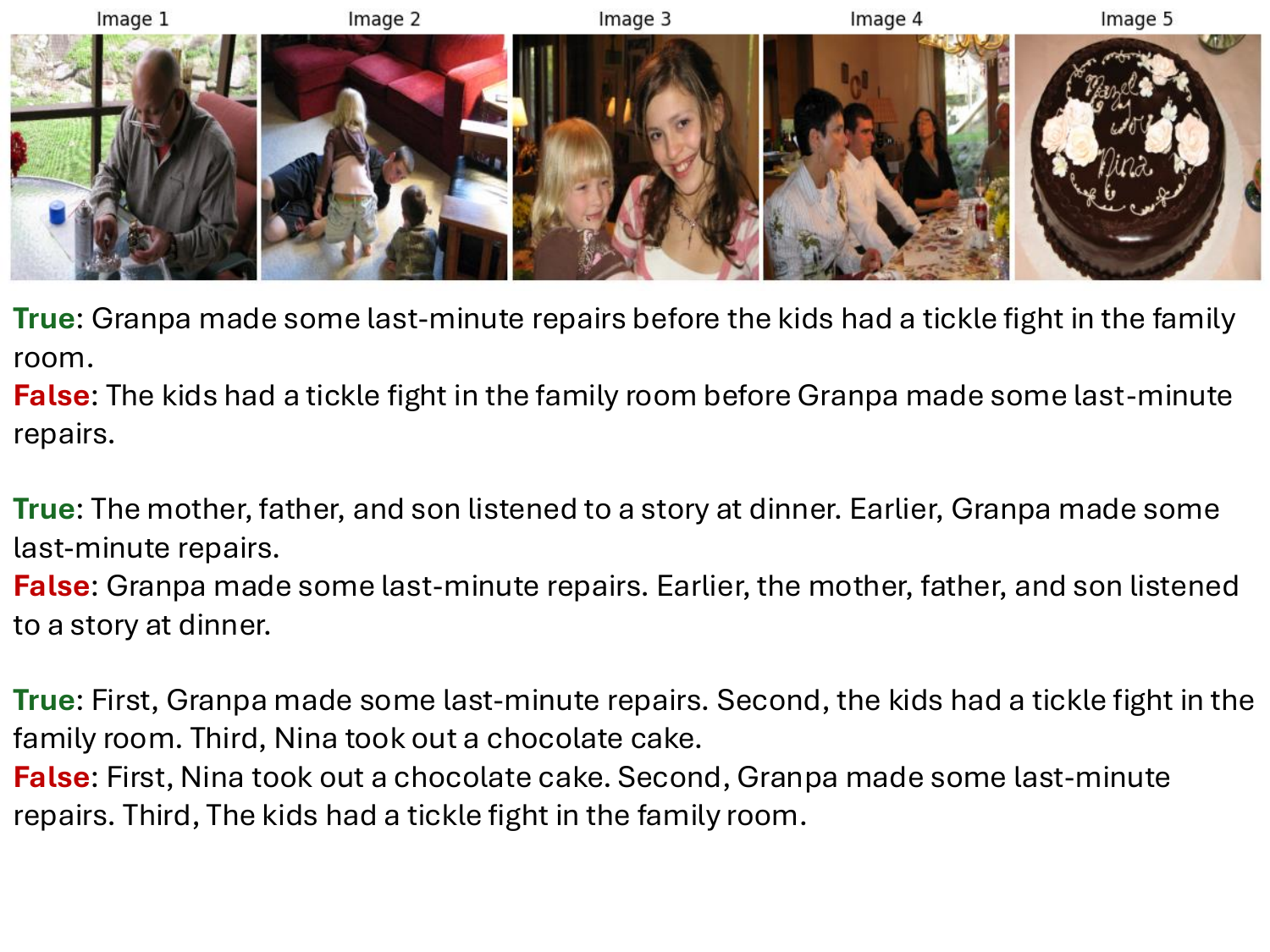} \\
    \includegraphics[width=\linewidth]{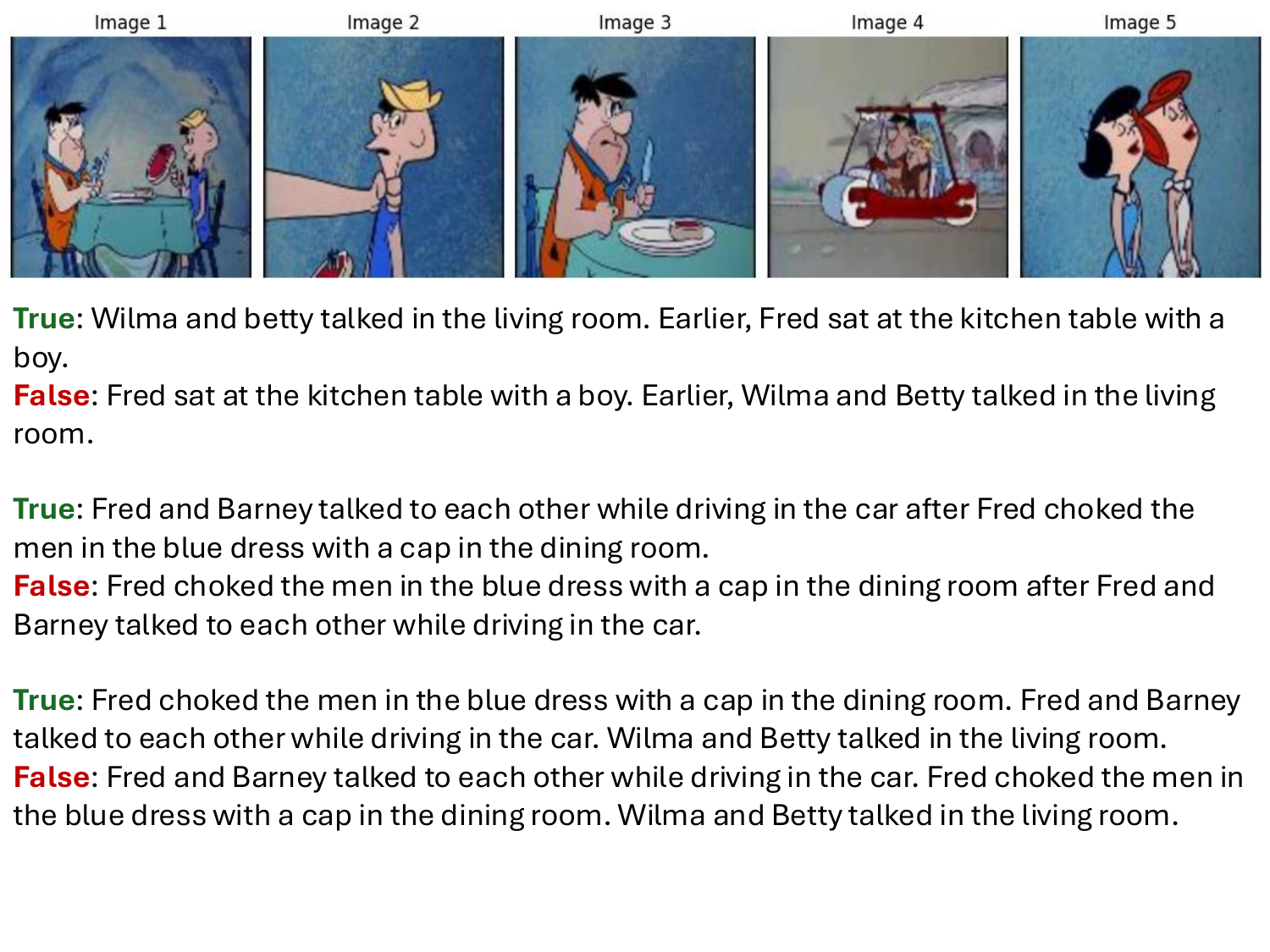} \\
\end{figure}

\end{document}